\pdfoutput=1

\documentclass[11pt]{article}

\usepackage[]{acl}
\usepackage{amsfonts} 

\usepackage{times}
\usepackage{latexsym}
\usepackage{graphicx}
\usepackage{multirow}

\usepackage[T1]{fontenc}

\usepackage{lipsum}  

\usepackage{amsmath}

\usepackage{tcolorbox}
\definecolor{mycolor}{RGB}{230, 245, 255}
\definecolor{orange}{RGB}{255, 240, 204}
\definecolor{colorgreen}{RGB}{240, 255, 220}

\usepackage{enumitem}

\usepackage{amsthm} 

\usepackage{xcolor} 

\definecolor{promptcolor}{RGB}{225, 255, 240} 

\usepackage[utf8]{inputenc}
\usepackage{booktabs}
\usepackage{microtype}

\title{Creating an AI Observer: Generative Semantic Workspaces} 

\author{Pavan Holur, Shreyas Rajesh, David Chong, Vwani Roychowdhury\\
         Department of Electrical and Computer Engineering, UCLA \\ 
         \texttt{\{pholur,shreyasrajesh38,davidchong13807,vwani\}@ucla.edu}}

\begin{document}
{\makeatletter\acl@finalcopytrue
  \maketitle
}
\begin{abstract}

An experienced human Observer reading a document --such as a crime report-- creates a succinct plot-like \textit{``Working Memory''} comprising different actors, their prototypical roles and states at any point, their evolution over time based on their interactions, and even a map of missing Semantic parts anticipating them in the future.  \textit{An equivalent AI Observer currently does not exist}. We introduce the \textbf{[G]}enerative \textbf{[S]}emantic \textbf{[W]}orkspace (GSW) -- comprising an \textit{``Operator''} and a \textit{``Reconciler''} -- that leverages advancements in LLMs to create a generative-style Semantic framework, as opposed to a traditionally predefined set of lexicon labels. Given a text segment $C_n$ that describes an ongoing situation, the \textit{Operator} instantiates actor-centric Semantic maps (termed ``Workspace instance'' $\mathcal{W}_n$). The \textit{Reconciler} resolves differences between $\mathcal{W}_n$ and a ``Working memory'' $\mathcal{M}_n^*$ to generate the updated $\mathcal{M}_{n+1}^*$. GSW outperforms well-known baselines on several tasks ($\sim 94\%$ vs. FST, GLEN, BertSRL - multi-sentence Semantics extraction, $\sim 15\%$ vs. NLI-BERT, $\sim 35\%$ vs. QA). By mirroring the real Observer, GSW provides the first step towards Spatial Computing assistants capable of understanding individual intentions and predicting future behavior.




\end{abstract}















\section{Overview}

\begin{figure}[t!]
    \centering
\includegraphics[width=\columnwidth]{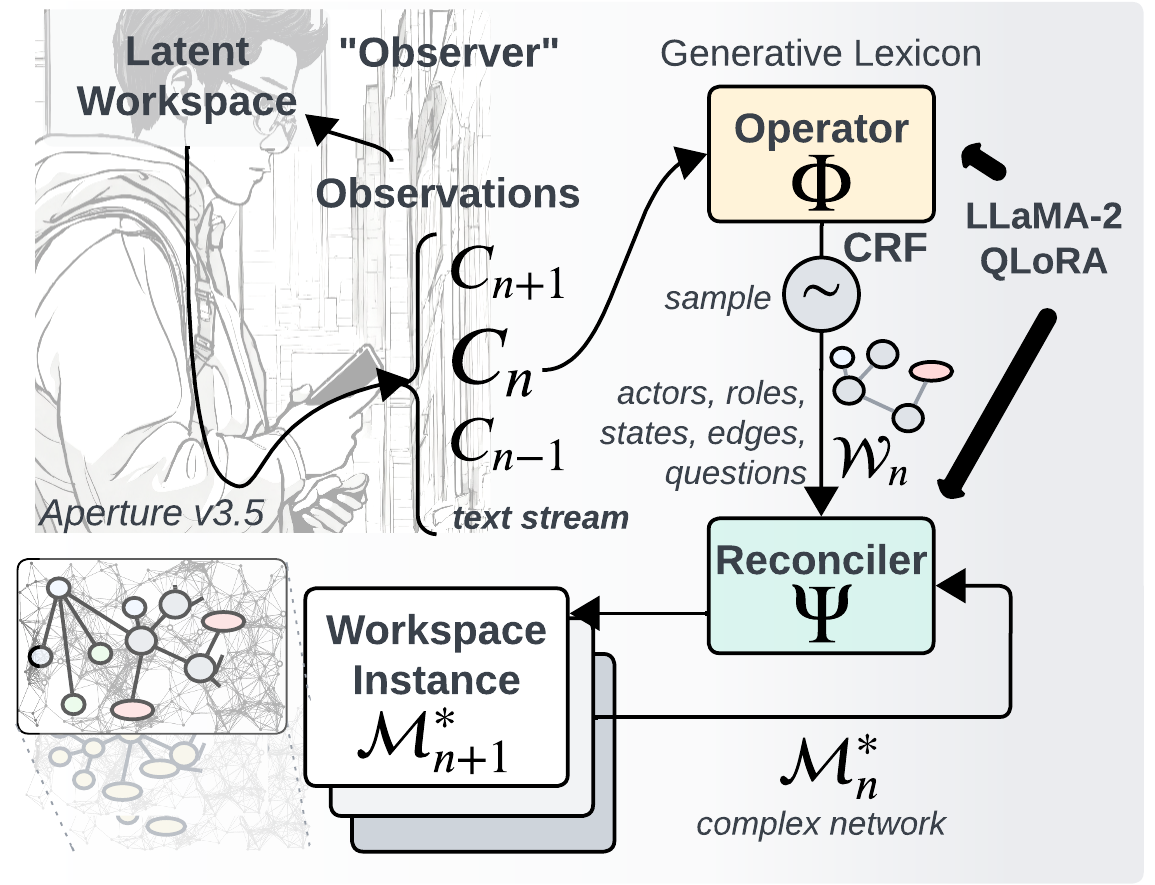}
    \caption{\textbf{The GSW framework:} The Observer constructs an internal map of the Semantics common across several instances of a situation, and in the process, identifies the recurring and prototypical conceptual gestalts. Any unfolding instance of a situation is processed through such semantic "lenses" and embedded in natural language with the use of grammar: a sample from the Observer's semantic map, with encodings of instance-specific actors, their interactions, and their evolution over space and time. In the GSW model, the goal is to construct an AI equivalent that interprets situations as they are encoded in text, to create a succinct \textit{``Workspace''} instance -- or working memory -- that contains an extendable layout of the Semantics (see Tab.~\ref{tab:operator-framework} for comparison to baselines). The Workspace is modeled as a Conditional Random Field (CRF), with Actors, Roles, States, Questions, and Predicates sampled from a conditional distribution. The CRF is estimated using a multi-task LLaMA + LoRA setup (\textit{Operator}). The \textit{Reconciler} computes the similarity between a pair of \textit{``Workspace instances''} sampled from the CRF (see Tab.~\ref{tab:all-reconciler} for a comparison to baselines).}
    \label{fig:overview}
\end{figure}



\begin{figure*}[t!]
    \centering
\includegraphics[trim={0cm 0 13cm 25cm},clip,width=\textwidth]{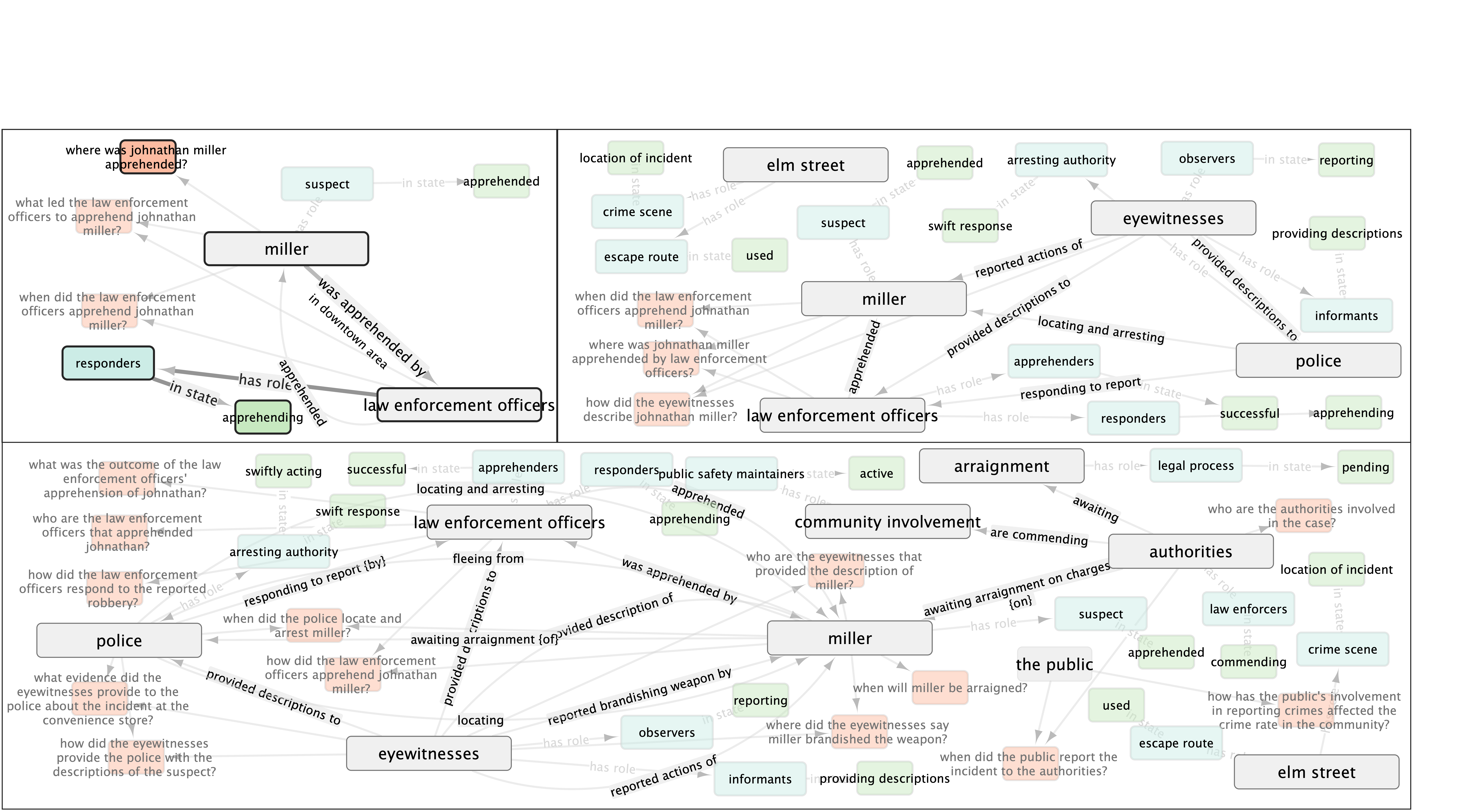}
    \caption{\textbf{Operator + Reconciler - Dynamics of the Generative Semantic Workspace [GSW]:} Snapshots of the evolving Workspace instance are depicted (from top-left, clockwise) along different time points, as the GSW framework processes a story $[C_1,\dots,C_N]$ (see Fig.~\ref{fig:cj0} for full story). We denote the lexicon types - Actor: grey, Role: blue, State: green, Predicate: edge, and Question: red - the first instance of each is in bold in the first pane. The workspace instances $\mathcal{W}_n$ (for each $C_n$) are aggregated into a consensus Workspace Instance $\mathcal{M}_{n+1}^*$ (see Fig.~\ref{fig:overview}) using the \textit{Reconciler} to compare the consensus Semantics at the previous step $\mathcal{M}_{n}^*$ with the latest Operator output $\mathcal{W}_n$. More workspace instances are presented in App. Sec.~\ref{app:sec:examples}.}
    \label{fig:spotlight}
\end{figure*}

We introduce a novel computational framework termed the \textbf{[G]}enerative \textbf{[S]}emantic \textbf{[W]}orkspace for estimating the latent Semantic structure in natural language from its embodiment in large-scale data using Large Language Models (LLM) (see Fig.~\ref{fig:overview}). GSW differentiates itself from existing methods~\cite{propbank, framenet, verbnet} by introducing an "Observer"-centric formalism (see Fig.~\ref{fig:overview}, Sec.~\ref{sec:study}). In addition to bringing in a novel computational aspect to the study of Semantics, such a framework would enable applications such as digital assistants in Spatial Computing (SC) and Augmented Reality (AR) interfaces, that require an Observer-centric view of the world.



Computationally, GSW comprises two parts: (a) The \textit{Operator}, a parameterized generative model of actor-centric Semantics as a Conditional Random Field (CRF) -- a sample instance is presented in Tab.~\ref{tab:first_sample}, and (b) The \textit{Reconciler}, a fine-tuned LLM model that compares and resolves a pair of Operator-generated instances, so that logical and temporal continuity of actor semantics is preserved (see Tab.~\ref{tab:subtasks}).




The rest of the paper is arranged as follows: In Sec.~\ref{sec:study}, we introduce an Observer-centric view of Semantics, App. Sec.~\ref{sec:relatedwork} contains related computational efforts, Sec.~\ref{sec:challenges} enumerates the two challenges in existing methods, Sec.~\ref{sec:approach} introduces the GSW formalisms, Secs.~\ref{sec:evalmethods}, Sec.~\ref{sec:results} presents results and discussion. FAQ is presented in App. Sec.~\ref{app:faq}, and GSW-generated Semantic maps are presented in App. Sec.~\ref{app:sec:examples}.




\section{A Cognitive Perspective for Observer-centric Semantics} \label{sec:study}

We begin by walking through the following post from an Observer's perspective to get a better sense of the Semantics critical to them:

\begin{center}
\label{post:A}
\noindent \textbf{[Sample S1]} 
\noindent Yesterday, in a swift response to a reported \textit{\textbf{robbery}}, \textbf{law enforcement officers} apprehended \textbf{Johnathan Miller}, a 32-year-old resident of \textbf{Greenview Avenue}, in the downtown area.
\end{center}

The Observer upon reading this first encounters the actors ``law enforcement officer'', ``johnathan miller'', ``greenview avenue'', ``robbery'', and, next comes up with \textit{``consensus''} identifiers that convey the underlying Semantics: Stereotypical \textbf{role} identifiers such as the \textit{Suspect}, \textit{Crime Scene}, and \textit{Apprehenders}, along with their respective \textbf{state} identifiers (e.g., "Where in the process is the suspect now?" $\rightarrow$ "apprehended"), delineate the responsibilities of the various \textbf{actors} in the situation. Role-states pre-condition the potential inter-actor relationships or \textit{predicates} ("responded to", "apprehended by"). Unanswered \textbf{questions} ("What led to...," "How did law enforcement...") act as placeholders for the missing Semantics parts that are currently unanswered in the post. A summary is presented in Tab.~\ref{tab:first_sample}. 

The goal of an AI Observer is to be able to generate such semantic maps. It is thus instructive to create a framework of how human Observers are able to create and understand such maps. 
An Observer's focus on Semantics is oriented toward the actors involved in a situation and the inter-actor dynamics, a fundamental principle encapsulated in \textit{"Situated Cognition"}~\cite{situated}. The Observer lives in an uncertain world with partial information forcing them to create a structured probabilistic model of potential outcomes via the roles, and states (constraining the inter-actor interactions) that the Observer attributes to the actors. \textit{Thus, a particular role, state, or interaction signifies a distribution} over ensuing outcomes, and cannot be described in entirety by few samples. This poses a communication challenge when two Observers --who have experienced the same situation -- have to communicate Semantic information over natural language. The Observers are forced to come up with a restricted set of consensus labels and syntax (or \textit{\textbf{Identifiers}}) that elicit similar distributions over the Semantics of the situation in both parties~\cite{kripke, jackendoff}. Thus, the usage of the word "suspect" intends to convey an underlying distribution of outcomes in terms of other labels or words -- such as accusation, freedom, penalties, and victimhood -- which themselves correspond to distributions. A Conditional Random Field (CRF)~\cite{crf} is one of many probabilistic modeling tools that can succinctly represent such scenarios.  If an Identifier cannot convey such a distribution over related concepts in a reliable way, then it is not an effective conveyor of Semantic information. These motivations are used to construct the GSW framework in Sec.~\ref{sec:approach} but first we review why some of the existing frameworks for constructing Semantic maps cannot be easily adapted for this task.

\begin{table}[t]
        \resizebox{\columnwidth}{!}{%
\begin{tabular}{@{}l|l@{}}
\toprule
\textbf{Semantic Parts} &
  \textbf{Identifiers} \\ \midrule
\textbf{\begin{tabular}[c]{@{}l@{}}ACTOR \\ specific named entities,\\ and other noun phrases\end{tabular}} &
  \begin{tabular}[c]{@{}l@{}}law enforcement officer, \\ Johnathan Miller, Greenview \\ Avenue, robbery\end{tabular} \\ \midrule
\textbf{\begin{tabular}[c]{@{}l@{}}ROLE \\ stereotypical purpose \\ embodied by each actor\end{tabular}} &
  \begin{tabular}[c]{@{}l@{}}apprehenders, suspect, \\ suspect residence, crime event\end{tabular} \\ \midrule
\textbf{\begin{tabular}[c]{@{}l@{}}PREDICATE \\ interactions between \\ actors in role-states\end{tabular}} &
  \begin{tabular}[c]{@{}l@{}}law enforcement officers -- responded \\ to -- reported robbery\\ {[}yesterday{]} \\ Johnathan Miller -- apprehended \\ by -- law enforcement officers \\ {[}in downtown area{]}\end{tabular} \\ \midrule
\textbf{\begin{tabular}[c]{@{}l@{}}STATE \\ attributes of actor-roles \\determined by context\end{tabular}} &
  \begin{tabular}[c]{@{}l@{}}successful, apprehended,\\ inhabited, reported\end{tabular} \\ \midrule
\textbf{\begin{tabular}[c]{@{}l@{}}QUESTION \\ unresolved valences \\ that need answering\end{tabular}} &
  \begin{tabular}[c]{@{}l@{}}What led to the apprehension of \\ Johnathan Miller by the law \\ enforcement officers?\\ How did the law enforcement officers \\ apprehend Johnathan Miller?\end{tabular} \\ \bottomrule
\end{tabular}}

    \caption{\textbf{An Observer's distribution of semantics are aliased to identifiers in NL:} Identifiers serve to imply the Observer's internal Semantics distribution of the roles and states played by the different actors as well as the predicates with which these actors interact. Missing facets that the Observer expects to have answered in the future, are represented as unresolved questions. Readers interpret these identifiers by reconstructing the original Semantic distributions. GSW is able to model such distributions and constructs extendable Semantic maps as a result (see Fig.~\ref{fig:spotlight}, App. Sec.~\ref{app:sec:examples}).}
    \label{tab:first_sample}
\end{table}






\section{Can existing frameworks construct Observer-centric Semantics?}

Over the last few decades, seminal works including PropBank, FrameNet, GLEN, ACE, and ConceptNet~\cite{propbank, framenet, glen, ace, conceptnet} have posited computational approximations to model the Semantic space and have addressed its modeling in two parts: (A) Constructing an explicit \textit{\textbf{Lexicon}} -- a set of pre-indexed \textit{Identifiers} that act as proxies for the underlying concept distributions, and (B) An accompanying \textbf{\textit{Text-to-Semantics }} model to map natural language (like \textbf{Sample S1}) onto the lexicon. For example, PropBank~\cite{propbank} \textit{enumerates} role identifiers that attempt to identify stereotypical (thematic) participants in the usage of a \textit{verb}; FrameNet~\cite{framenet} groups semantic role identifiers into larger groups known as Frames and adds Lexical Units (LU) -- mostly verbs and nouns -- that map a Frame to natural language. Frame Elements (FE) correspond to the roles. The Text-to-Semantics model is a post-hoc model that is rule-based or data-driven model~\cite{fst, srlbert, text2event}, which maps individual text segments onto this lexicon. The study of lexicons and Text-to-Semantics models has a long and rich history that is captured in great detail in the Related Works (see Appendix Sec.~\ref{sec:relatedwork}). 


\subsection{Challenges} \label{sec:challenges}
 
Consider \textbf{Sample S1} mapped onto the FrameNet~\cite{framenet} lexicon using the Frame Semantic Transformer (FST)~\cite{fst}, a Text-to-Semantics model  (Tab.~\ref{Tab:FST_sit})\footnote{Similar comparisons to Event and SRL frameworks have been performed at scale in Sec.~\ref{sec:operator_results}}. In this setup, predicates (verbs and many nouns) behave as anchors that map to a specific subset of enumerated frame identifiers -- such as \textit{Response}, \textit{Statement}, \textit{Robbery}, \textit{Arrest}, \textit{Residence} -- and text spans are associated to frame elements (FE) that are part of each frame. For example, the ``Authorities'' FE is the identifier representing a semantic role in the \textit{Arrest} frame and corresponds to the ``law enforcement officers'' span. A thorough analysis reveals the following properties:

\begin{table}[ht!]
\addtolength{\tabcolsep}{-2pt}
\resizebox{\columnwidth}{!}{%
\small
\begingroup
\renewcommand{\arraystretch}{0.6} 
\begin{tabular}{l|l|l}
\toprule
\textbf{Frame} & \textbf{FE} & \textbf{Span} \\ \midrule
\multicolumn{1}{l|}{\textbf{Response}} & \multicolumn{1}{l|}{Manner} & swift \\ \cmidrule(l){2-3} 
\multicolumn{1}{l|}{\textbf{}} & \multicolumn{1}{l|}{Trigger} & to a reported robbery \\ \midrule
\multicolumn{1}{l|}{\textbf{Statement}} & \multicolumn{1}{l|}{Message} & Robbery \\ \midrule
\multicolumn{1}{l|}{\textbf{Robbery}} & \multicolumn{1}{l|}{\cellcolor[HTML]{EFEFEF}} & \cellcolor[HTML]{EFEFEF} \\ \midrule
\multicolumn{1}{l|}{\textbf{Arrest}} & \multicolumn{1}{l|}{Authorities} & law enforcement officers \\ \cmidrule(l){2-3} 
\multicolumn{1}{l|}{\textbf{}} & \multicolumn{1}{l|}{Suspect} & \begin{tabular}[l]{@{}l@{}}Johnathan Miller, a 32-year-old \\ resident of greenview avenue\end{tabular} \\ \midrule
\multicolumn{1}{l|}{\textbf{Residence}} & \multicolumn{1}{l|}{Resident} & Johnathan Miller \\ \cmidrule(l){2-3} 
\multicolumn{1}{l|}{\textbf{}} & \multicolumn{1}{l|}{Location} & of greenview avenue \\
\bottomrule
\end{tabular}
\endgroup
}
\caption{\textbf{Frame Semantic Transformer applied on Sample S1:} The retrieved frames, frame elements and text spans are presented above. These are compared to the Semantics identified by GSW in Tab.~\ref{tab:operator-framework}.}
\label{Tab:FST_sit}
\end{table}


\noindent \textbf{[L1] Fragmented Semantics:} While the FrameNet lexicon identified (rows in Tab.~\ref{tab:first_sample}) from \textbf{Sample S1} are meaningful at the frame-level, challenges arise when trying to ascertain the Semantics associated to a specific actor and their inter-actor relationships encoded in \textbf{Sample S1}. For instance, the \textit{Response} frame neglects to address "Who responded?", the ``Responding Entity'' FE. Next, although "what the \textit{Responders} responded to?" is mapped to the phrase "to a reported robbery" (FE "Trigger"), the frame ``Robbery'' is without any frame elements, though "Johnathan Miller" is a candidate for the "Perpetrator" FE, thus missing a vital connection between the \textit{Response} and \textit{Robbery} frames. Additionally, the \textit{Statement} frame is misplaced -- the lexical unit ``reported'' is being taken out-of-context -- adding noise to the Semantic map since it's treated equally alongside more contextually pertinent frames. 





\noindent \textbf{[L2] Discontinuous Semantics:} Fragmented Semantics at the sentence level exacerbates the challenges when trying to construct a dynamic semantic map, where actors, their roles, states, and their interactions evolve according to a consensus plot-like situation model. Consider for example, the result of applying FST on \textbf{Sample S2} which follows \textbf{Sample S1} (see Fig.~\ref{fig:cj0} for the full story):

\begin{center}
\label{post:B}
\textbf{[Sample S2]} 
\textbf{Police} swiftly acted on the provided descriptions, locating and arresting \textbf{Miller} within the hour.
\end{center}

The resulting semantics from FST are as follows: \textbf{Frame 1:} \textit{Intentionally\_act} $\rightarrow$ \textit{Agent: police}, \textit{Manner: swiftly}, \textit{means: on the provided descriptions}, \textbf{Frame 2:} \textit{Locating} $\rightarrow$ \textit{Perceiver: Police}, \textit{Sought entity: Miller}, \textbf{Frame 3:} \textit{Arrest} $\rightarrow$ \textit{Authorities: police}, \textit{Suspect: Miller}, \textit{Time: within the hour}. Establishing connections \textit{between} the semantics from \textbf{Sample S1} to \textbf{Sample S2} requires connecting the dots better: Are the two \textit{Arrest} frames identified in the adjacent samples referencing the same \textit{arrest}? Are "\textit{Police}" and "\textit{Law Enforcement}" entities the same \textit{authority} FE? The latest extraction lacks information about why \textit{Miller} was arrested; there's no mention of the \textit{Robbery} identified in the previous frame. We now propose an alternate Semantics framework that does not inherit these challenges.

\section{The Generative Semantic Workspace} \label{sec:approach}

To motivate our approach, we first describe its underlying intuitive human-like workflow. Assume the Observer has encountered a \textbf{\textit{situation}} $S$, and is reading $X_S$ -- the unrolling of a situation instance in $S$ given by a sequence of text segments: $[C_1, \dots, C_N] \in X_S$. For example, the Observer reads $C_1=$ \textbf{Sample S1} and then reads $C_2=$ \textbf{Sample S2}, and so on. As in Tab.~\ref{tab:first_sample}, the Observer generates a \textbf{\textit{Workspace instance}} $\mathcal{W}_1$ which comprises the actors, their roles, states, inter-actor relationships and unresolved questions. A network representation of $\mathcal{W}_1$ is shown in Fig.~\ref{fig:spotlight}. Clearly we need a computational model -- that we call the \textbf{\textit{``Operator''}} -- to codify this step of going from a text segment to the Workspace instance. 

Next, the Observer reads \textbf{Sample S2} and generates another Workspace instance $\mathcal{W}_2$ \textit{involving the same person}, \textit{Miller} and crime, \textit{robbery}. Clearly, the actors may have acquired new roles or states. For example, the question \textit{"What led the law enforcement to apprehend johnathan miller?"} is unresolved given \textbf{Sample S1} and shows up as a valence node in $\mathcal{W}_1$ (see Fig.~\ref{fig:spotlight} (left, top)). However, upon reading \textbf{Sample S2}, we notice that the question is answered: "\textit{... acted on the provided descriptions ...}". As a result, the question is now outdated and must be removed from the workspace instance (see Fig.~\ref{fig:spotlight}, right top). This requires another computational unit -- \textbf{\textit{Reconciler}} -- that takes the relevant parts of $\mathcal{W}_1$ and $\mathcal{W}_2$ and merges them into a \textbf{\textit{Working memory}} instance $\mathcal{M}_2$ which will be then updated to create $\mathcal{M}_3$ by reconciling it with $\mathcal{W}_3$ and so on. 

To formalize this intuitive model of the Observer, we denote the family of all possible workspace instances, including all possible working memory instances as the Workspace $\mathbb{W}$ (inspired by the concept of Global Workspace Theory~\cite{baars}). A subspace specific to situation $S$ is denoted $\mathbb{W}_S$. Then the \textbf{Operator} can be formally defined as $f_{\Phi_S}: \mathcal{X}_S \rightarrow \mathbb{W}_S$, and the \textbf{Reconciler}, $g_{\Psi_S}: \mathbb{W}_S \times \mathbb{W}_S \rightarrow \mathbb{W}_S$. Appendix \ref{app:sec:llm} explains how LLMs (i.e. $\Phi_S,\Psi_S$) make the dynamically created  labels in workspace instances  semantically consistent; that is, semantically similar context would lead to similar semantic labels.

\subsection{Training the Operator model}

\begin{figure}[h!]
    \centering
\includegraphics[width=0.80\columnwidth]{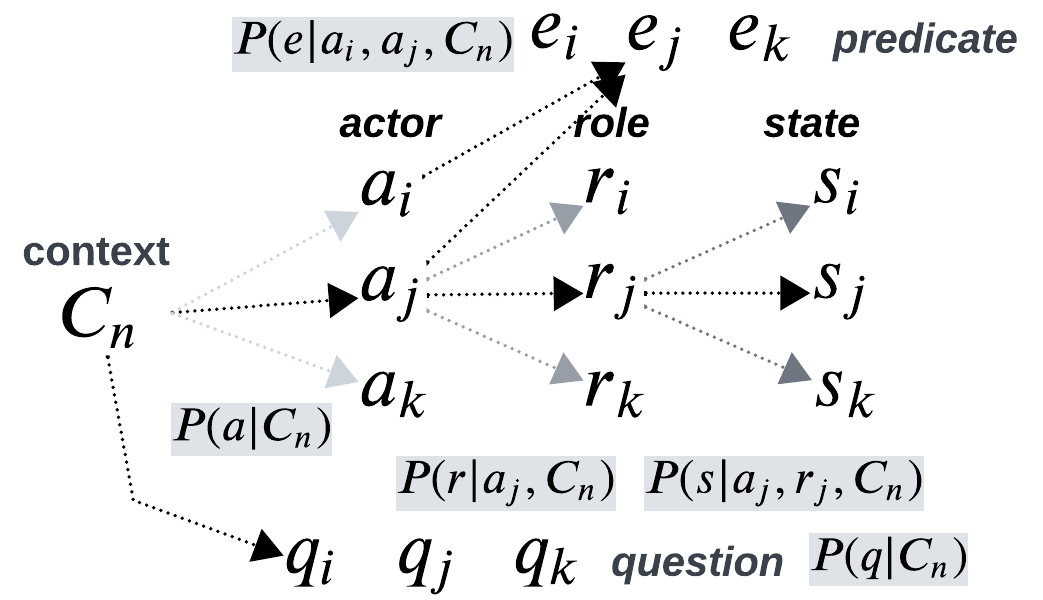}
    \caption{\textbf{Conditional Random Field model of the Workspace:} A workspace instance is constructed by sampling the CRF given an input text segment $C_n$.}
    \label{fig:crf}
\end{figure}

Given an input sample $C_n$, such as \textbf{Sample S1},  the Operator model outputs a set of roles, states, and predicates for each actor that describe the inter-actor relationships, and questions about the Semantics that are yet to be answered going forward (see~Tab.\ref{tab:first_sample} and Fig.~\ref{fig:spotlight}). Formally, $f_{\Phi}$ is trained to output a Workspace instance $W_n$ as sampled from a parameterized distribution across the workspace $\mathbb{W}$: i.e. $P_\mathbb{W}(\mathcal{W}_n | C_n; \Phi)$. Observer-centric Semantics discussed in Sec.~\ref{sec:approach} intuits that this distribution is best expressed as a CRF, where a set of Identifiers sampled from a Semantics distribution are in fact \textit{themselves} representative of conditional distributions that lead to the generation of more granular Identifiers. In our setup, the CRF is comprised of conditional distributions estimated for each Semantic part from Tab.~\ref{tab:first_sample} -- see Fig.~\ref{fig:crf}. A sampled path through this CRF results in a Workspace instance.

\subsection{Workspace instance as a Complex Network: A probabilistic interpretation}

One can view the task of jointly estimating a workspace instance as the estimation of a \textit{complex network}:
$$\mathcal{W}:=\mathcal{W}(V,E),$$ 
where the nodes $v \in V$ and edges $e \in E$ correspond to the following lexicon map:

\noindent \textbf{Node ($v$):} \{Actor $a$, Role $r$, State $s$\}; \{Question $q$\}\\
\noindent \textbf{Edge ($e$):} Source Node $v_S$ $\rightarrow$ Predicate $\rightarrow$ Target Node $v_T$.

\noindent Then, the Workspace instance itself can be viewed as a sample drawn from the Workspace:
\begin{align*}
\mathcal{W} \sim f_{\Phi}(C_n) \implies V_{\mathcal{W}}, E_{\mathcal{W}} \sim f_{\Phi}(C_n).
\end{align*}
Assume conditional independence: 
$$P(V_{\mathcal{W}}, E_{\mathcal{W}} | C_n) \approx \prod P(\{v_S, e, v_T\} | C_n).$$ 
\begin{align*}
P(\{v_S, e&, v_T\} | C_n) = \\&P(e | v_S, v_T, C_n) \times P(v_S, v_T | C_n).
\end{align*}
Now, $P(v_S, v_T | C_i) = P(v_S| C_n) \times P(v_T| C_n)$. Each can be further decomposed:
\begin{align*}
P(v | C_n) &:= P(\{a, r, s\} | C_n) \\
&= P(s | a, r; C_n) \\ &  \times P(r | a; C_n) \times P(a | C_n).
\end{align*}

\subsection{Reconciler: Comparing Workspace Instances as Complex Networks}

\begin{table}[h]
\small
\centering
\setlength{\tabcolsep}{0.30em} 
\begin{tabular}{@{}ll@{}}
\toprule
\multicolumn{2}{c}{\textbf{Reconciler Classification Labels}}                                                                                                                                                                                                                                    \\ \midrule
\multicolumn{1}{l|}{\begin{tabular}[c]{@{}l@{}}Nodes, \\Edges\end{tabular}} & \begin{tabular}[c]{@{}l@{}}0: the existing workspace instance is sufficient\\ 1: the new instance overwrites the workspace\\ 2: both instances are important and unrelated\end{tabular} \\ \midrule
\multicolumn{1}{l|}{\begin{tabular}[c]{@{}l@{}}Questions\end{tabular}}      & \begin{tabular}[c]{@{}l@{}}0: question remains unanswered and relevant\\ 1: question is answered or irrelevant\end{tabular}                                                             \\ \bottomrule
\end{tabular}
\caption{\textbf{Summary of classification tasks trained in the Reconciler:} Nodes and edges are classified pairwise. \textit{Reconciliation} or REC applies on pairwise nodes and edges (0: keep previous, 1: replace, 2: keep both), and \{question, node\} and \{question, edge\} pairs are classified -- \textit{Question Resolution} or QR -- as answered or irrelevant (1) or unanswered and relevant (0).}
\label{tab:reconciler-labels}
\end{table}

\vspace{1em}

Consider two Workspace instances $\{\mathcal{W}_i, \mathcal{W}_{i+1}\} \in \mathbb{W}$: the Reconciler compares the two instances to construct an aggregated workspace instance $\mathcal{M} \in \mathbb{W}$. Since we represent the instances as actor-centric Complex Networks -- $\mathcal{W}_i := \mathcal{W}_i (V,E)$, $\mathcal{W}_{i+1} := \mathcal{W}_{i+1} (V,E)$, 
\begin{align*}
 \mathcal{M} & \sim g_{\Psi}(\mathcal{W}_i, \mathcal{W}_{i+1}) \\
\iff          \langle V_{M}, E_{M} \rangle & \sim g_{\Psi}(\langle V_{i}, E_{i} \rangle, \langle V_{i+1}, E_{i+1} \rangle),
\end{align*}
the Reconciler can be formulated as a classification task, where every pair of nodes and edges are compared as follows: 
\begin{align*}
d & = g_{\Psi}(v_{i}, v_{i+1}), g_{\Psi}(e_{i}, e_{i+1}).
\end{align*}
A summary of the possible decisions is presented in Tab.~\ref{tab:reconciler-labels}: (i) Retain the node/edge currently in $\mathcal{W}_i$ and discard the node/edge in $\mathcal{W}_{i+1}$ (as noise or infeasible); (ii) Discard the node/edge in $W_i$ and incorporate the node/edge in $\mathcal{W}_{i+1}$; (iii) Keep both $\mathcal{W}_i, \mathcal{W}_{i+1}$ nodes/edges. Questions are compared with both nodes and edges. 

Despite the Reconciler being posed as a \textit{local} node/edge-level model that doesn't consider the two Workspace instances in their entirety, the actor-centric nature allows modifications at a smaller neighborhood-level: \textit{an actor's associated Semantics can only be influenced by the actors around them.} Empirically, such a Reconciler demonstrates high performance at aggregating Workspace instances: see classification performance in Tab.~\ref{tab:all-reconciler} and networks in App. Sec.~\ref{app:sec:examples}.

\noindent \textbf{Actor-Centric Memory and Autoregressive Semantics}: With the Operator and Reconciler trained, they can be paired in an auto-regressive setup: given a sequence of consecutive text segments: $\{C_1, \dots, C_N\} \in \mathcal{X}$, the Operator outputs Workspace instances $\{\mathcal{W}_1, \dots, \mathcal{W}_N\} \in \mathbb{W}$, and the Reconciler is deployed in an auto-regressive fashion to continually update a consensus Workspace instance $\mathcal{M}^*$ such that:
$$
\mathcal{M}_{n+1}^* \leftarrow g_\Psi(\mathcal{M}_{n}^*, \mathcal{W}_n).
$$


\subsection{Practical Considerations}

\subsubsection{Data}

News reports are a popular resource for sampling semantics-rich stories that belong to universally recognized situation patterns (in an effort to cater to a wider readership). We query GDELT~\cite{gdelt}, a Jigsaw-powered news-indexing platform, with Situation identifiers to retrieve a small set of situation-conditioned \texttt{en\_US} articles.  Tab.~\ref{tab:data} presents statistics about the data. These situations were manually selected as an initial seed set -- similar to FrameNet's early versions containing few frames~\cite{fillmore} -- to assess the validity of the GSW framework (see FAQ~\ref{sec:extensions} for more on our choice of situations).

\begin{table}[h]
\centering
\small\addtolength{\tabcolsep}{-3pt}
\begingroup
\renewcommand{\arraystretch}{0.4} 
\begin{tabular}{@{}c|c|c|c@{}}
\toprule
\textbf{Situation Label}              & \textbf{Documents} & \textbf{Sentences} & \textbf{Tokens}  \\ \midrule
\textit{crime and justice}      & 80  & 1209  & 100,635 \\ \midrule
\textit{fire fighting}          & 79   & 1116  & 87,901  \\ \midrule
\textit{technology development} & 81  & 1334  & 122,493 \\ \midrule
\textit{healthcare}             & 81  & 1259  & 117,962 \\ \midrule
\textit{economy}                & 78   & 1264   & 110,605  \\ \bottomrule
\end{tabular}
\endgroup
\caption{\textbf{Data Statistics:} Situation-specific news reports are sampled from GDELT. Each document (or article) is split into short contexts $C_1, \dots, C_N$ (of $3$ sentences). Just \textasciitilde 80 articles per situation is sufficient to finetune LoRA adapters to project \textit{any} text segment onto the Semantic workspace.  
}
\label{tab:data}
\end{table}

\subsubsection{Silver-Standard Annotations and Computational Requirements} 

\begin{figure}[h]
    \centering
\includegraphics[width=0.8\columnwidth]{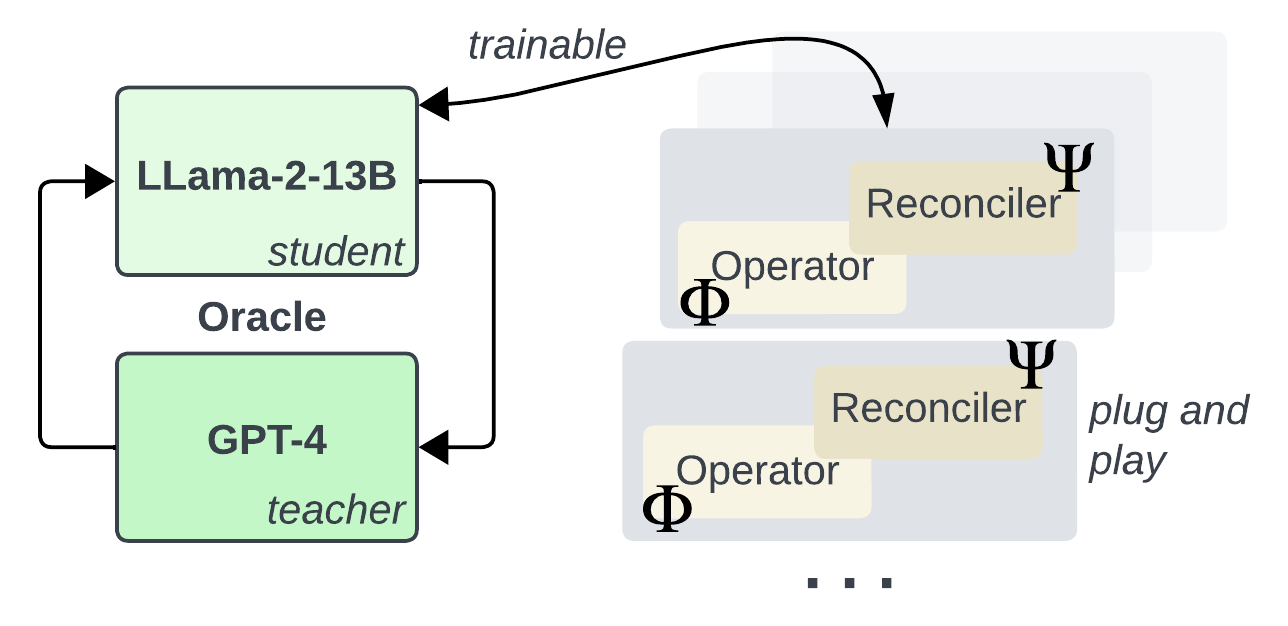}
    \caption{\textbf{Modular parameter sets for the Workspace model:} Every Operator and Reconciler model (specific to a situation) comprises $<1M$ parameters -- using PeFT and LoRA -- and relies on the same shared LLM Oracle (LLaMA-2-13B). }
    \label{fig:distributed}
\end{figure}

GPT-4~\cite{gpt4} is found to be excellent at generating Workspace instances by using carefully~\cite{prompt} constructed multi-turn prompts (see Codebase) facilitated by Guidance\footnote{ \small \url{www.github.com/guidance-ai/guidance}} and given short contexts $\dots, C_i, \dots, C_j, \dots$ (3 sentence window) drawn from GDELT news articles. However, smaller, open-source models such as LLaMA~\cite{llama2} and Mistral~\cite{mistral} models (including Chat-pretrained) struggle to output the semantics without finetuning. In order to construct open-source, hostable, but still powerful approximations of GPT-4 Operators locally, adapters such as LoRA~\cite{lora} are now available as data- and parameter-efficient tunable modules (see PeFT~\cite{prefixtuning,llamaadapter}). That is:
$$
f_{\hat\Phi} (C_i, \mathcal{T}) \sim f_\Phi (C_i, \mathcal{T}).
$$
Different QLoRA~\cite{qlora} adapters are learned for each situation and can be inserted into a shared LLaMA backbone to match situation-specific input contexts (plug-and-play) (see Fig.~\ref{fig:distributed}). Each adapter only contains ~1M trainable parameters (rank-2) and requires a surprisingly few number of data samples to be finetuned (see Tab.~\ref{tab:data}). FAQ~\ref{sec:computational} details our training setup.

\section{Evaluation Methods} \label{sec:evalmethods}

We evaluate the Operator and Reconciler across multiple situations listed in Tab.~\ref{tab:data} using human evaluation via UpWork service as well as existing model baselines detailed below. An ablation study across situations and model architectures is provided in the Appendix (see Sec. ~\ref{sec:ablation}, ~\ref{sec:extend}, ~\ref{fig:cross}).

\noindent \textbf{Operator:} The workspace instances $\mathcal{W}_i$ generated from $C_i$ are evaluated with respect to baselines by UpWork workers (also see App. Sec.~\ref{app:annotator}): \textbf{[UpWork Task 1]} Perform a side-by-side blind evaluation with existing workspace instance generations output by state-of-the-art (SOTA) workspace model \textit{baselines}: (i) Frame Semantic Transformer (FST)~\cite{fst}, (ii) BERT for Semantic Role Labeling (BertSRL)~\cite{srlbert}, and an (iii) Event Extraction (GLEN)~\cite{glen};  \textbf{[UpWork Task 2]} Given questions generated by the Operator and GPT-4 from the same corpus, perform a blind rating of question quality on a scale from 1(worst) - 5(best).

\begin{table}[h]
\centering
\small
\begingroup
\renewcommand{\arraystretch}{0.4} 
\begin{tabular}{@{}l|l|l@{}}
\toprule
\textbf{Task}                     & \textbf{Baseline}       & \textbf{Reconciler Labels}                  \\ \midrule
\multirow{3}{*}{\textbf{REC/NLI}} & Entailment     & 0 (Keep old)                \\ \cmidrule(l){2-3} 
                         & Contradiction  & 1 (Replace)                 \\ \cmidrule(l){2-3} 
                         & Neutral        & 2 (Keep old and new)        \\ \midrule
\multirow{2}{*}{\textbf{QR/QA}} & 0 (Unanswered) & 0 (Unanswered \& relevant) \\ \cmidrule(l){2-3} 
                         & 1 (Answered)   & 1 (Answered or irrelevant)  \\ \bottomrule
\end{tabular}
\endgroup
\caption{\textbf{An approximate mapping from Baselines to Reconciler tasks:} Pre-existing NLI baseline models can be repurposed as a Reconciler and similarly a Question-Answering model can, to an extent, perform Question Reconciliation.}
\label{tab:NLI_QA_mapping}
\end{table}

\vspace{1em}

\noindent \textbf{Reconciler:} We identify 2 tasks and baselines for the Reconciliation and Question Resolution tasks: [\textbf{B1}] \textit{Natural Language Inference (NLI):} Given a pair of ordered text segments -- the \textit{premise} and the \textit{hypothesis}, output either (a) \textit{entailment:} the hypothesis is entailed by the premise; (b) \textit{contradiction:} the hypothesis contradicts the premise; or (c) \textit{neutral:} the hypothesis and premise can co-exist. A suitable mapping of labels from the Reconciliation task to NLI is shown in Tab.~\ref{tab:NLI_QA_mapping}. Models: RoBERTa~\cite{roberta} and DeBERTa~\cite{deberta} fine-tuned on the SNLI~\cite{snli} dataset. [\textbf{B2}] \textit{Question-Answering:} Given a \textit{context, question} tuple, output an answer as a span within the input context. A question is \textit{resolved} if QA returns a non-empty span. Models: RoBERTa~\cite{roberta} fine-tuned on SQuAD~\cite{squad}.

\section{Results and Discussion} \label{sec:results}

We present qualitative and quantitative results for the Operator and Reconciler models below. Several Workspace instances created as a result of parsing short stories are presented in App. Sec.~\ref{app:sec:examples}.

\begin{table}

\addtolength{\tabcolsep}{-3.5pt}
\begingroup
\renewcommand{\arraystretch}{0.4} 
\resizebox{\columnwidth}{!}{%
\begin{tabular}{@{}cccc@{}}
\toprule
\multicolumn{4}{c}{\textbf{Ours} (Baseline)}                                                                                    \\ \midrule
\multicolumn{1}{c|}{\textbf{Situation}} &
  \multicolumn{1}{c|}{\begin{tabular}[c]{@{}c@{}}vs. Zhan et al\\ (GLEN) \end{tabular}} &
  \multicolumn{1}{c|}{\begin{tabular}[c]{@{}c@{}}vs. Shi \& Lin\\ (BERT-SRL)\end{tabular}} &
  \begin{tabular}[c]{@{}c@{}}vs. Chanin\\ (FST)\end{tabular} \\ \midrule
\multicolumn{1}{c|}{\textit{crime \& justice}} & \multicolumn{1}{c|}{\textbf{0.90} (0.10)} & \multicolumn{1}{c|}{\textbf{0.96} (0.04)} & \textbf{0.7} (0.3)   \\ \midrule
\multicolumn{1}{c|}{\textit{economy}}          & \multicolumn{1}{c|}{\textbf{0.98} (0.02)} & \multicolumn{1}{c|}{\textbf{0.96} (0.04)} & \textbf{0.86} (0.14) \\ \midrule
\multicolumn{1}{c|}{\textit{firefighting}}     & \multicolumn{1}{c|}{\textbf{0.98} (0.02)} & \multicolumn{1}{c|}{\textbf{0.98} (0.02)} & \textbf{0.79} (0.21) \\ \midrule
\multicolumn{1}{c|}{\textit{healthcare}}       & \multicolumn{1}{c|}{\textbf{1.0} (0.00)}  & \multicolumn{1}{c|}{\textbf{0.96} (0.04)} & \textbf{0.94} (0.06) \\ \midrule
\multicolumn{1}{c|}{\textit{tech. dev.}}       & \multicolumn{1}{c|}{\textbf{0.96} (0.04)} & \multicolumn{1}{c|}{\textbf{0.96} (0.04)} & \textbf{0.86} (0.14) \\ \bottomrule
\end{tabular}%
}
\endgroup
\caption{\textbf{Operator Evaluations - Part 1 - Comparison with Existing Frameworks:} Given a short context, English-speaking annotators are shown the unlabeled outputs of the Operator and a baseline framework (GLEN, BERT-SRL, FST) and asked to select the one which best summarizes the semantics in the text. The Operator is preferred over baselines across situations.}
\label{tab:operator-framework}
\end{table}

\vspace{2em}

\begin{table}[h]
\small
\begingroup
\renewcommand{\arraystretch}{0.4} 
\begin{tabular}{@{}c|c|c|c@{}}
\toprule
\textbf{Situation} & \textbf{GPT-4 (d*)} & \textbf{Operator (d)} & \textbf{| d - d* |} \\ \midrule
\textit{crime \& justice} & 4.45 +/- 0.98 & 4.47 +/- 0.81 & = 0.02 \\ \midrule
\textit{economy} & 4.84 +/- 0.50 & 4.74 +/- 0.72 & = 0.10 \\ \midrule
\textit{firefighting} & 4.80 +/- 0.66 & 4.63 +/- 0.91 & = 0.17 \\ \midrule
\textit{healthcare} & 4.88 +/- 0.47 & 4.53 +/- 1.14 & = 0.35 \\ \midrule
\textit{tech. dev.} & 4.84 +/- 0.67 & 4.61 +/- 0.94 & = 0.23 \\ \midrule
\multicolumn{1}{c|}{{\color[HTML]{656565} (Avg.: 4.68)}} & \multicolumn{1}{c|}{{\color[HTML]{656565} 4.76}} & {\color[HTML]{656565} 4.60} &  \\ \bottomrule
\end{tabular}
\endgroup
\caption{\textbf{Operator Evaluations - Part 2 - Quality of Generated Questions:} English-speaking annotators are asked to rate the quality of Operator-generated questions given a short context $C_i$ on a scale from 1(worst)-5(best). Questions generated by GPT-4 for the \textit{same} contexts are used as a baseline. The  Operator model is observed to generate questions of similar quality to GPT-4.}
\label{tab:operator-questions}
\end{table}

\begin{table*}
\centering
\small\addtolength{\tabcolsep}{+2pt}

\begingroup
\renewcommand{\arraystretch}{0.4} 

\begin{tabular}{@{}c|ccc|cccc@{}}
\toprule
                                & \multicolumn{3}{c|}{\textbf{ACC.} $\uparrow$}                                                                                  & \multicolumn{4}{c}{\textbf{F1} $\uparrow$}                                                                                                                                            \\ \midrule  \textbf{Situation} & \multicolumn{7}{c}{\textbf{Reconciliation}} \\ \midrule
                       & \multicolumn{1}{c|}{He et al.} & \multicolumn{1}{c|}{Liu et al.} & \textbf{Reconciler} & \multicolumn{1}{c|}{He et al. } & \multicolumn{1}{c|}{Liu et al. } & \multicolumn{1}{c|}{\textbf{Reconciler}} & \textit{\textbf{Sensitivity}} \\ \midrule
\textit{crime and justice}      & \multicolumn{1}{c|}{0.72}                         & \multicolumn{1}{c|}{0.71}                       & \textbf{0.81} & \multicolumn{1}{c|}{0.69}                         & \multicolumn{1}{c|}{0.68}                       & \multicolumn{1}{c|}{\textbf{0.81}} & \textit{\textbf{0.91}}          \\ \midrule
\textit{fire fighting}          & \multicolumn{1}{c|}{0.71}                         & \multicolumn{1}{c|}{0.69}                       & \textbf{0.89} & \multicolumn{1}{c|}{0.68}                         & \multicolumn{1}{c|}{0.66}                       & \multicolumn{1}{c|}{\textbf{0.89}} & \textit{\textbf{0.99}}          \\ \midrule
\textit{technology development} & \multicolumn{1}{c|}{0.74}                         & \multicolumn{1}{c|}{0.69}                       & \textbf{0.87} & \multicolumn{1}{c|}{0.71}                         & \multicolumn{1}{c|}{0.61}                       & \multicolumn{1}{c|}{\textbf{0.87}} & \textit{\textbf{0.92}}          \\ \midrule
\textit{healthcare}             & \multicolumn{1}{c|}{0.74}                         & \multicolumn{1}{c|}{0.70}                       & \textbf{0.91} & \multicolumn{1}{c|}{0.71}                         & \multicolumn{1}{c|}{0.65}                       & \multicolumn{1}{c|}{\textbf{0.91}} & \textit{\textbf{0.96}}          \\ \midrule
\textit{economy}                & \multicolumn{1}{c|}{0.76}                         & \multicolumn{1}{c|}{0.70}                       & \textbf{0.88} & \multicolumn{1}{c|}{0.74}                         & \multicolumn{1}{c|}{0.63}                       & \multicolumn{1}{c|}{\textbf{0.88}} & \textit{\textbf{0.96}}   \\    \midrule     & \multicolumn{7}{c}{\textbf{Question Resolution}} \\ \midrule
                       & \multicolumn{1}{c|}{\begin{tabular}[c]{@{}c@{}}Liu et al. \\ (Base)\end{tabular}} & \multicolumn{1}{c|}{\begin{tabular}[c]{@{}c@{}}Liu et al. \\ (Large)\end{tabular}} & \textbf{Reconciler} & \multicolumn{1}{c|}{\begin{tabular}[c]{@{}c@{}}Liu et al. \\ (Base)\end{tabular}} & \multicolumn{1}{c|}{\begin{tabular}[c]{@{}c@{}}Liu et al. \\ (Large)\end{tabular}} & \multicolumn{1}{c|}{\textbf{Reconciler}} & \textit{\textbf{-}}  \\ \midrule

\textit{crime and justice}      & \multicolumn{1}{c|}{0.53}                         & \multicolumn{1}{c|}{0.59}                       & \textbf{0.92} & \multicolumn{1}{c|}{0.49}                         & \multicolumn{1}{c|}{0.57}                       & \multicolumn{1}{c|}{\textbf{0.92}} &  -                              \\ \midrule
\textit{fire fighting}          & \multicolumn{1}{c|}{0.63}                         & \multicolumn{1}{c|}{0.62}                       & \textbf{0.97} & \multicolumn{1}{c|}{0.60}                         & \multicolumn{1}{c|}{0.60}                       & \multicolumn{1}{c|}{\textbf{0.98}} &  -                              \\ \midrule
\textit{technology development} & \multicolumn{1}{c|}{0.74}                         & \multicolumn{1}{c|}{0.67}                       & \textbf{0.99} & \multicolumn{1}{c|}{0.72}                         & \multicolumn{1}{c|}{0.64}                       & \multicolumn{1}{c|}{\textbf{0.99}} &  -                              \\ \midrule
\textit{healthcare}             & \multicolumn{1}{c|}{0.69}                         & \multicolumn{1}{c|}{0.68}                       & \textbf{0.97} & \multicolumn{1}{c|}{0.67}                         & \multicolumn{1}{c|}{0.66}                       & \multicolumn{1}{c|}{\textbf{0.97}} &  -                              \\ \midrule
\textit{economy}                & \multicolumn{1}{c|}{0.63}                         & \multicolumn{1}{c|}{0.64}                       & \textbf{0.94} & \multicolumn{1}{c|}{0.60}                         & \multicolumn{1}{c|}{0.60}                       & \multicolumn{1}{c|}{\textbf{0.94}} &  -                              \\ \bottomrule
\end{tabular}%

\endgroup
\caption{\textbf{Reconciler performance on different situations across Reconciliation and Question Resolution Tasks and (pre-existing) Baselines:} The two (sub-)tasks -- Reconciliation [REC] (3 classes), and Question Resolution [QR] (2 classes) -- are described in Tab.~\ref{tab:subtasks}. Baselines for the two subtasks include SOTA NLI~\cite{snli} and QA~\cite{roberta} pipelines -- see Sec~\ref{sec:evalmethods} -- that are existing tasks that share similar goals. As seen across situations, the Reconciler outperforms the baselines handily (REC - $>25\%$, QR - $>50\%$). }
\label{tab:all-reconciler}
\end{table*}

\subsection{Operator} \label{sec:operator_results}
We present the comparison with workspace instance baselines in Tab.~\ref{tab:operator-framework}. Operator outputs on a held-out test set are strongly preferred ($ >90\% $ outputs selected) over 
GLEN~\cite{glen}, BertSRL~\cite{srlbert} and Frame Semantic Transformer~\cite{fst} (>70\%) for all situations. A comparison of Operator-generated questions with GPT-4 is presented in Tab.~\ref{tab:operator-questions}. Operator question quality is highly rated across all situations ($4.6$ average rating) and is close to the GPT-4 quality for each situation ($<0.35$ avg. rating difference).

\subsection{Reconciler}


The classification results of the Reconciler are presented in Tab.~\ref{tab:all-reconciler}. The LLaMA + LoRA Reconciler outperforms the DeBERTa~\cite{deberta} and RoBERTa~\cite{roberta} baselines by a wide margin with an improvement of $ >25\% $ at the task of Reconciliation and $ > 50\%$ at Question Resolution on Accuracy (ACC) and F1 score (weighted) on a held-out test set. Reconciliation is a 3-class classification task and Question Resolution is a binary classification task; the test set is made sure to be class-balanced. \textbf{Sensitivity} (see Tab.~\ref{tab:all-reconciler} - right) merges Class 1 (replace) and Class 2 (keep both) in the Reconciliation task. A high Sensitivity score indicates that the Reconciler model is capable of correctly deciding whether information from a new Operator instance is to be added to the consensus workspace instance $\mathcal{M}^*$ or ignored.

\section{Concluding Remarks}



An Observer-centric semantic representation scheme is an essential evolutionary construct that has allowed species to predict and dynamically adapt to an uncertain and constantly changing world, and the proposed computational approximation --\textit{Generative Semantic Workspace}--promises to add similar abilities to NLU frameworks. GSW demonstrates considerable success at constructing consensus Workspace instances -- colloquially referred to as ``Working memory'' -- in which actor-centric Semantics are aggregated (by the \textit{``Operator''}) and modified continually (by the \textit{``Reconciler''}) given a stream of text segments. Evaluations using human annotators and baseline NLU models have revealed that the resulting Workspace instances convey a plot-like Semantic summary that is more comprehensive than its counterparts. Like a true Working memory, the consensus workspace instances generated by GSW, incrementally and in a compositional manner, stitch together the salient Semantic features of actors -- the roles, states -- and the inter-actor relationships (predicates). Unanswered questions provide a means of looking to the future to complete the missing valences of the past.

\begin{figure}[h!]
    \centering
    \includegraphics[width=0.90\columnwidth]{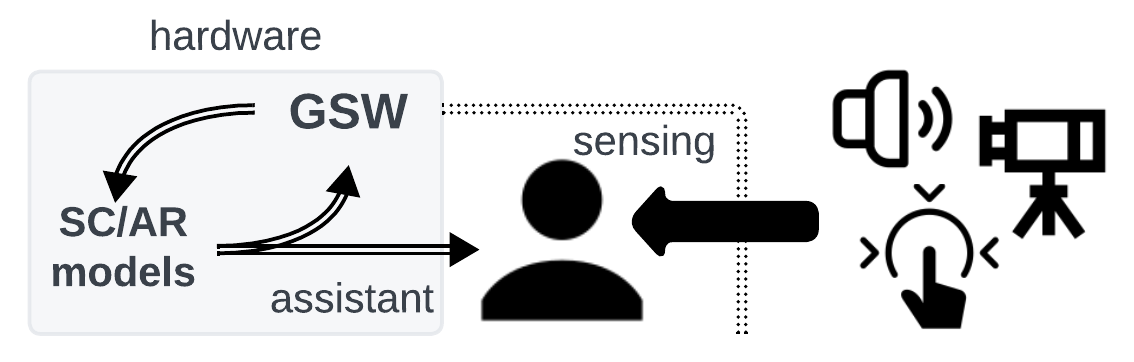}
    \caption{\textit{A prospective use-case of GSW - Targeted information retrieval of Semantics in SC/AR settings.}}
    \label{fig:conclusion}
\end{figure}

\vspace{1em}

Among the several prospective real-world applications, the Complex Network representation of the Workspace instance may facilitate multi-hop actor-lexicon \textit{traversals} -- using such existing KG-QA methods as DeepPath~\cite{deep_path} -- that can help uncover unrivaled, long-range Semantic connections across the Workspace: a novel use-case within Targeted Information Retrieval (or \textit{Semantic Search}). More broadly, GSW can be envisioned to form the basis of a foundation model for augmented reality (AR) or spatial computing (SC) frameworks: a \textit{real} Observer and a corresponding AI assistant encounter experiences in concert, which in turn enables the AI assistant to facilitate reasoning about Observer intentions and thus, predict, and even perhaps, \textit{preempt} decision-making. 

\newpage

\section{Limitations}

\subsection{Technical challenges in the GSW implementation} 

[\textit{Coreference resolution}] Workspace instances require long-range coreference resolution to map entity mentions across multiple sentences to a shared actor (that constitutes a node). This module can at times pose inaccuracies. [\textit{Data variety}] Training and testing of the Operator and Reconciler models are conducted on news media reports which are cleaner, linear, and structured when compared to other text datasets such as social media posts. [\textit{Retrieval in the Reconciler:}] As the Workspace instance grows larger, the Reconciler has to infer more pairwise decisions between nodes and edges in the latest Operator output $\mathcal{W}_n$ and the consensus Workspace instance $\mathcal{M}_n^*$. We have implemented several heuristics to reduce the number of comparisons: for example, using actor mentions in $\mathcal{W}_n$ to identify a substantially smaller subgraph $\tilde{\mathcal{M}}_n^* \subset \mathcal{M}_n^*$ that is likely to be updated. Such heuristics may result in some Semantic parts becoming outdated if the subgraph $\tilde{\mathcal{M}}_n^*$ does not include all the nodes and edges that need updates. [\textit{Recall vs. Precision:}] The Operator is trained with a carefully balanced negative sampling hyperparameter ($40\%$), to avoid the generation of spurious lexicon samples -- observed frequently in the generation of predicates -- that comprise the Workspace instance: training data for negative samples are also retrieved from querying GPT-4 (see Codebase). As a result, the occasional input context may yield too many or too few Semantic parts.



\subsection{Potential Risks and Ethical Considerations} 
Given that GSW uses a LLaMA-backbone, the same potential risks and ethical considerations apply as with any pretrained generative-style language model: that the Operator model (CRF of lexicon distributions) can reflect the bias in the finetuning dataset and as a result, can construct potentially harmful, violent, racist and sexist lexicon samples. But the bias can be more nuanced: roles attributed to actors (noun phrases) that are described in the training data in the same stereotypical roles, may be once again associated with the actor during inference \textit{despite the input context suggesting an entirely different role}. Our experiments found very little evidence of this form of bias.

\bibliographystyle{acl_natbib}
\bibliography{anthology,custom}

\appendix

\section{Related Computational Models of Workspaces} \label{sec:relatedwork}

Existing workspace modeling efforts have depended, to a large extent, on manually-annotated lexicon ontologies. Among the most popular efforts are PropBank~\cite{propbank} and FrameNet~\cite{framenet}, which attempt to define a correspondence between (a) the syntactic ``realizations'' of the semantics \textit{explicit} within the language structure of a sentence, and (b) a finite and discrete set of semantic ``roles''~\cite{levin}. Both methods used expert linguists to define and annotate the role set and the role definitions. We detail the similarities and differences below:


\noindent \textbf{PropBank:} PropBank utilized a \textit{bottom-up} approach: (1) Dependency Parse Trees~\cite{dpt} were applied to a large text corpus to distill shared syntactic patterns (``Framesets'') specific to each verb (a process known as ``lexical sampling''). (2) For each Frameset, the corresponding sentences were manually annotated with an enumerated set of \textit{arguments} \texttt{ARG:0,\dots, ARG:N}. These arguments were later associated to verb-specific definitions using VerbNet~\cite{verbnet}. The  semantic roles are identified as corresponding \textit{spans} within the sentence (commonly a \texttt{NP}, \texttt{NNP} subtree in the dependency parsing). For example, the sentence (A):
    \begin{center}

    \noindent \textbf{Officers} captured \textbf{Sarah} at the \\ \textbf{Sepulveda on-ramp} of the \textbf{405}.
    \end{center}
would be annotated with the arguments: 
\begin{center}

\noindent \textit{Agent}: \textbf{officers}, \textit{Predicate}: \textbf{captured}, \textit{Patient}: \textbf{Sarah}.
\end{center}
Perhaps the greatest benefit of PropBank was that its syntactic ``grounding'' made it possible for rule-based and early ML models~\cite{rml, srlbert} to \textit{learn the task of distilling the semantics} given a sentence, albeit within the confines of a \textit{ limited} ontology of $>3000$ verbs and $>4000$ Framesets 

\noindent \textit{Event Databases:} PropBank evolved in several directions, including efforts to unify it with related semantic lexicon such as VerbNet and FrameNet~\cite{semlink, semlink2}, or augment it via the DWD overlay~\cite{dwdoverlay} to WikiData~\cite{wikidata}. The latter of these efforts now manifests as ``Event'' databases~\cite{eventreview, xiang2019survey} such as the ACE~\cite{ace} and ERE~\cite{ere} datasets, and led to the DARPA initiative of Event identification/extraction challenges. Events are best motivated by their related identification tasks: Given a sentence, identify the event(s) -- from a set of \textit{hundreds} of events in a pre-annotated schema~\cite{glen, eventextract, text2event} -- that the sentence is referring to. For example, (A) would be annotated with the \textit{Capture} event.

\noindent \textbf{FrameNet:} In contrast to PropBank and related Event ontologies, FrameNet\footnote{\url{https://framenet.icsi.berkeley.edu/}} utilizes a \textit{top-down} approach that is not tethered to the syntax structure. Rather, expert linguists aggregated roles (redefined as ``Frame Elements'' (FE)) from a large corpus of sentences, which are \textit{known to co-exist} under a conceptual gestalt, or ``Frame''. Each frame additionally comprises a set of ``Lexical Units'' (LU) - valences (mostly verbs and nouns) whose occurrence in a sentence increases the likelihood of a frame. For example,
\begin{center}
the \textit{Frame}: \textbf{Taking Captive} 
\end{center}
would contain the following frame elements and lexical units: 
\begin{center}
FE: \textit{Agent}, \textit{Captive}, \textit{Cause} \\
LU: \textit{capture.v}, \textit{secure.v}
\end{center}

FrameNet ($1000$s of frames and $10,000$s of FE) is a substantially larger and more comprehensive ontology~\cite{framenetcompare} compared to Propbank. When originally constructed, automated systems could not effectively identify the frames implied by a sentence; today, however, Transformer models~\cite{attention, bert, fst} have demonstrated success at accurately modeling the sentence-to-frame mapping.

Despite the enormous success and wide adoption of PropBank, FrameNet, and their descendant works, the explicit, finite, and discrete lexicons they employ raise the question: \textit{When is an explicit lexicon ontology complete?}.  While FrameNet provides Frame-Frame precedence and subset relationships, these are coarse-grained and do not adequately answer the question: \textit{How can we track the evolution of semantics across a stream of sentences?} - a key requirement for workspace models.

More recent work ~\cite{conceptnet} has attempted to assemble a comparatively larger (and less stringent) open-schema semantic ontology of concepts using game-play based crowd-sourcing techniques~\cite{verbosity}. However, such efforts to scale manual annotation ultimately do not address how a complete ontology can be constructed. Event Graph Models (EGM)~\cite{tegm} generate event networks to describe the dynamics of events in a text corpus, often using a combination of submodules such as Coreference Resolution~\cite{coref}, Named Entity Recognition (NER)~\cite{ner} and Semantic Role Labeling~\cite{srl}. Extensions~\cite{future} generate the \textit{most likely} event \textit{template} sequences. These methods rely on predefined event schema to enumerate the set of possible events. However, while EGMs both track the evolution of semantics across sentences and offer an unsupervised approach to extending existing ontologies, these often marginalize across individual contexts in the training corpus, and generate the most likely event \textit{schema} that follows a current event schema network. As a result, these works have yet to design methods to track the semantics across a specific document.

\section{Frequently Asked Questions}
\label{app:faq}

\subsection{Why LLMs for Operator and Reconciler models?} \label{app:sec:llm}

A Workspace instance $\mathcal{W}_n$ is generated as a sample from the Operator $f_\Phi$ given an input $C_n$ and the Reconciler $g_\Psi$ compares a pair of Workspace instances $\{\mathcal{W}_n, \mathcal{W}_{n+1}\}$. In this setup, there is no explicit guarantee of stability. 


There are two requirements to guarantee stability: (1) \textit{Operator:} Given a pair of text contexts $\{C_i,C_j\} \in \mathcal{X}$ that are similar in semantic meaning but differ in syntax and grammar, the corresponding Workspace instances $\{\mathcal{W}_i,\mathcal{W}_j\}$ must have similar Semantic parts -- actors, roles, states, predicates and questions -- that are inferred. I.e.:
\begin{align*}
f_{\Phi}& (C_i, \mathcal{T}) \sim  f_{\Phi} (C_j,\mathcal{T}) \\
& \forall \, \{C_i, C_j\} \in \mathcal{X}, \texttt{sem}(C_i) \approx \texttt{sem}(C_j),
\end{align*}
where $\mathcal{T}$ is the Semantics extraction task (specified via prompting or finetuning); (2) \textit{Reconciler:} Given two workspace instances $\{\mathcal{W}_i,\mathcal{W}_j\}$ that are similar in meaning but use different Identifiers, the Reconciler must be able to infer that these Identifiers refers to very similar Semantics distributions. I.e.:
\begin{align*}
g_{\Psi_S}& (\mathcal{W}_i, \mathcal{T}) \sim  g_{\Psi_S} (\mathcal{W}_j,\mathcal{T}) \\
& \forall \, \{\mathcal{W}_i, \mathcal{W}_j\} \in \mathbb{W}_S, \texttt{sem}(\mathcal{W}_i) \approx \texttt{sem}(\mathcal{W}_j).
\end{align*}
\noindent \textit{Role of Large Language Models:} Prior applications of LLMs -- such as T5~\cite{T5}, XLNet~\cite{xlnet}, MPlug~\cite{mplug}, FST~\cite{fst}, BERTopic~\cite{bertopic} -- have effectively identified a distilled set of semantic labels -- including topics, sentiments, and frames -- across text segments $\dots, C_i, \dots, C_j, \dots$ with similar meaning, despite significant variations in syntax and phrasing. Given its focus on identifying a distilled Semantics maps, GSW similarly employs LLMs (LLaMA specifically) to model the Operator and Reconciler and demonstrates through several experiments, that the instances are stable for short stories.

\begin{table}[ht!]
\small
\centering
\setlength{\tabcolsep}{1.3em} 
\begingroup
\renewcommand{\arraystretch}{0.4} 
\begin{tabular}{@{}ll@{}}
\toprule
\multicolumn{2}{c}{\textbf{Summary of Subtasks}}                                                                                                                                                                         \\ \midrule
\multicolumn{1}{l|}{\begin{tabular}[c]{@{}l@{}}Operator\\ (Generation)\end{tabular}}       & \begin{tabular}[c]{@{}l@{}}Actor / Role / State Identification,\\  Question and Predicate Generation\end{tabular} \\ \midrule
\multicolumn{1}{l|}{\begin{tabular}[c]{@{}l@{}}Reconciler\\ (Classification)\end{tabular}} & \begin{tabular}[c]{@{}l@{}}Node and Edge Reconciliation, \\ Question Resolution\end{tabular}                \\ \bottomrule
\end{tabular}
\endgroup
\caption{\textbf{Summary of subtasks addressed by the Operator and Reconciler:} The Operator model generates a Workspace instance $\mathcal{W}_n$ containing roles, states, predicates and questions -- by sampling a Conditional Random Field of Semantics -- described in Sec.~\ref{sec:study} --  given an input text segment $C_n$. The Reconciler compares Workspace instances (a consensus \textit{``Working memory''} $\mathcal{M}_n^*$ instance vs. latest $\mathcal{W}_n$) and outputs an updated consensus instance $\mathcal{M}_{n+1}^*$.}
\label{tab:subtasks}
\end{table}

\subsection{Why these situations?} \label{sec:extensions}

The selected situations for our study -- \textit{crime and justice, firefighting, technology development, healthcare, and the economy} -- were chosen for their frequent coverage in news, diverse roles, states, and predicates and varying durations of events. For instance, events that take place in a \textit{Crime and Justice} situation range from the act of committing a crime all the way up to the delivery of justice, typically unfolding linearly over shorter timescales compared to, for example, the lengthy, cyclical process of \textit{Technology Development}.

Furthermore, each situation can be roughly mapped to an entire set of conceptual gestalts -- such as frames and events -- that prior lexicon banks have annotated: for example, Crime and Justice can be associated to over $25$ frames in FrameNet, 15 for Firefighting, 35 for Technology Development, 50 for Healthcare, and over 60 for the Economy. Frames are manually annotated conceptual gestalts that are identified from large text corpuses (see Sec.~\ref{sec:relatedwork}). Therefore, a high coverage across the set of frames indicates that the corresponding Operator and Reconciler models are applicable to a wide range of situation instances. Furthermore, this mapping from situations to frames facilitates a fair comparison between GSW and frame-based Semantic frameworks. A similar correspondence is observed between situations and Event databases.


\subsection{How do we extend the number of Situations?} \label{sec:extend}

Currently our situations act as predefined (i.e., not dynamically generated by GSW)  conditional variables in the CRF model used for modeling Observer-centric semantics. Clearly, an overall awareness of the situation an observer is in influences the roles/states/interactions of the actors. There is, however, sufficient evidence in how humans and other species operate where the awareness of a situation itself can be predicted based on even higher-level cues. Based on these intuitions, we plan to pursue multiple avenues for making the AI observer more of a generalist and also enable it to adaptively create new situation labels based on experiences that do not sufficiently coincide with predictions based on existing situations based models. We articulate some of these approaches in the following. 

First, our ablation studies indicate that learning distributions over the Semantics across one situation is easily able to adapt to unfamiliar situations and generate and reconcile Workspace instances that rival the situation-focused model~\ref{tab:ablation-table}. In future research we will use a fewer number of LoRA units, where each unit will focus on multiple situations, thereby merging different situations into a single composite label. Merging can also be conducted at the model level as recent efforts such as TIES~\cite{ties} have demonstrated. This would result in a hierarchy much along the lines of how GDELT organizes its news articles into news topics and event databases are arranged into subclasses.


\subsection{What are the computational requirements for training and inferencing?} \label{sec:computational}

LoRA units are augmented to a LLaMA backbone and are finetuned over $10$ epochs ($\sim 72$ hours training per model), batch size of $8$, rank $2$ (all layers and heads, totalling $1M$ parameters), dropout $0.05$, weight scaling-parameter $\alpha=32$, maximum window length $w=1024$. We note that the use of LLaMA is freely available for research and requires a custom commercial license. Accordingly, our LoRA adapters + LLaMA models are also available under the same license.

Operator-to-Reconciler inference is performed sequentially (see Fig.~\ref{fig:overview}) and the corresponding LoRA weights can be off-loaded and on-loaded onto the shared LLaMA pretrained backbone to save time cost. The Operator model processes multiple contexts $\dots, C_i, \dots, C_j, \dots$ in batched mode, generating Workspace instances in parallel. Within each instance generation, since the lexicon CRF requires conditional generation, the roles corresponding to each actor are generated prior to generating the corresponding states and so forth.

\subsection{How is spatial and temporal information encoded in the Workspace?} \label{sec:spaceandtime}

Existing lexicon frameworks such as PropBank and FrameNet encode the spatial and temporal information through explicitly defined lexicon Identifiers. \color{black} In PropBank, verbs are annotated with a \textit{temporal argument} \texttt{[ARG-TMP]}, which, when using BertSRL as the Text-to-Semantics model, maps to text spans such as ``last week'', ``tomorrow'', ``at 10AM''. Similarly \texttt{[ARG-LOC]} qualifies spans that convey location (local, city, state, country geographical identifiers). In FrameNet, corresponding frames (such as \texttt{Temporal Pattern}, \texttt{Location in Time}) and frame-specific frame elements (\texttt{location/place/region}, \texttt{time}) are temporal and spatial information-sensitive lexicon that Text-to-Semantics models such as Frame Semantic Transformers attach to the key text spans.

The Global Semantics Workspace has its own mechanisms to capture spatial and temporal information. \\ \textbf{Capturing Spatial Information:} At the sentence level, the Operator encodes spatial information as actors (constituting nodes in our complex network representation), often with accompanying roles and states that provide context-specific information.

\begin{figure*}[ht]
    \centering
\includegraphics[trim={20cm 0 20cm 0},clip,width=\textwidth]{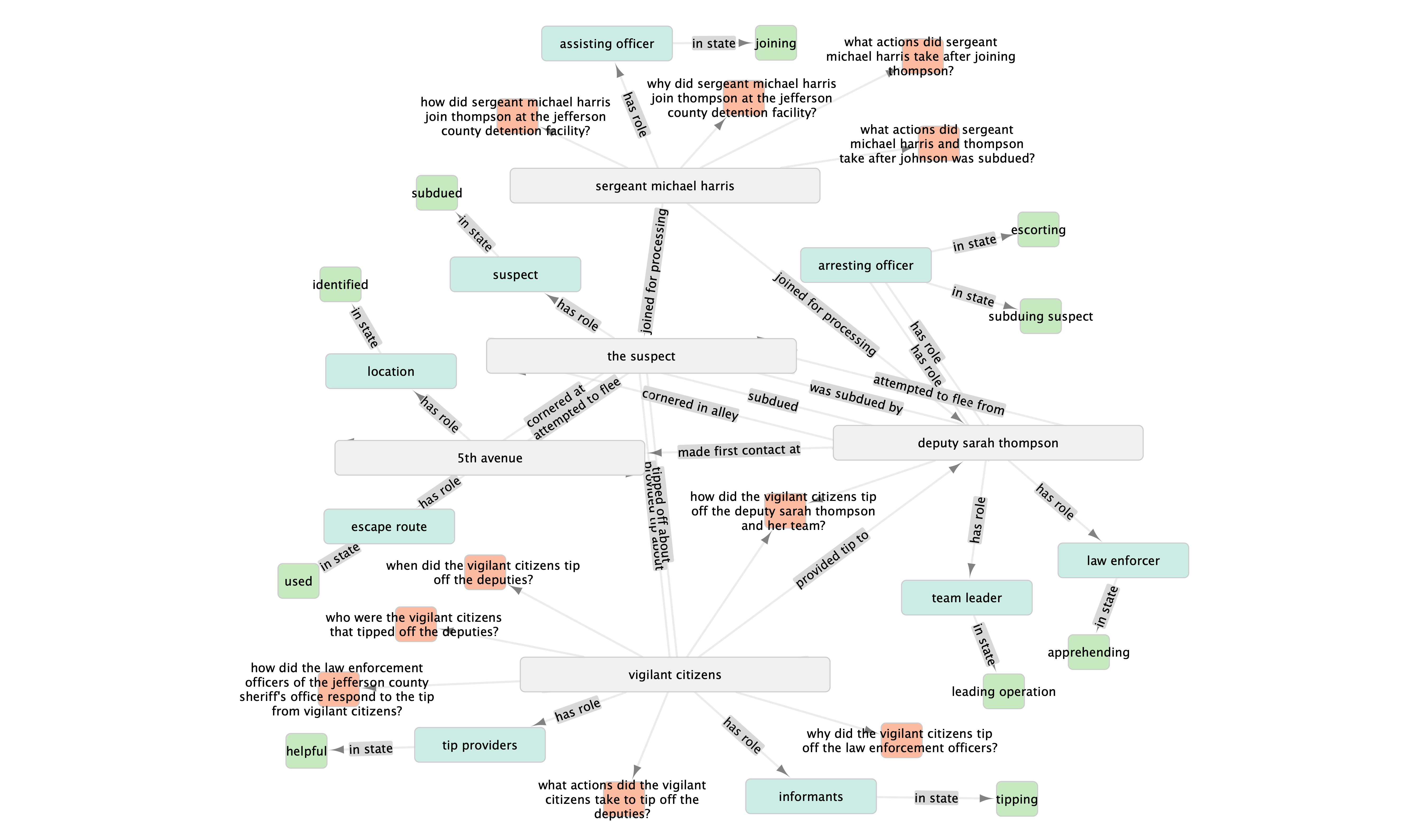}

    \caption{\textbf{Crime and Justice -} \textit{In a swift operation conducted by the law enforcement officers of the Jefferson County Sheriff's Office on Tuesday, the notorious fugitive, Mark Johnson, was apprehended in the downtown area of Birmingham, Alabama. Acting on a tip from vigilant citizens, Deputy Sarah Thompson and her team made first contact with the suspect, cornering Johnson in an alley off 5th Avenue South. Johnson was handcuffed as he attempted to flee. After being subdued, Thompson escorted Johnson to the patrol car, in which Sergeant Michael Harris joined him to the Jefferson County Detention Facility for processing.}}
    \label{fig:cj1}
\end{figure*}

\begin{figure*}[ht]
    \centering
\includegraphics[trim={0cm 0 0cm 0},clip,width=\textwidth]{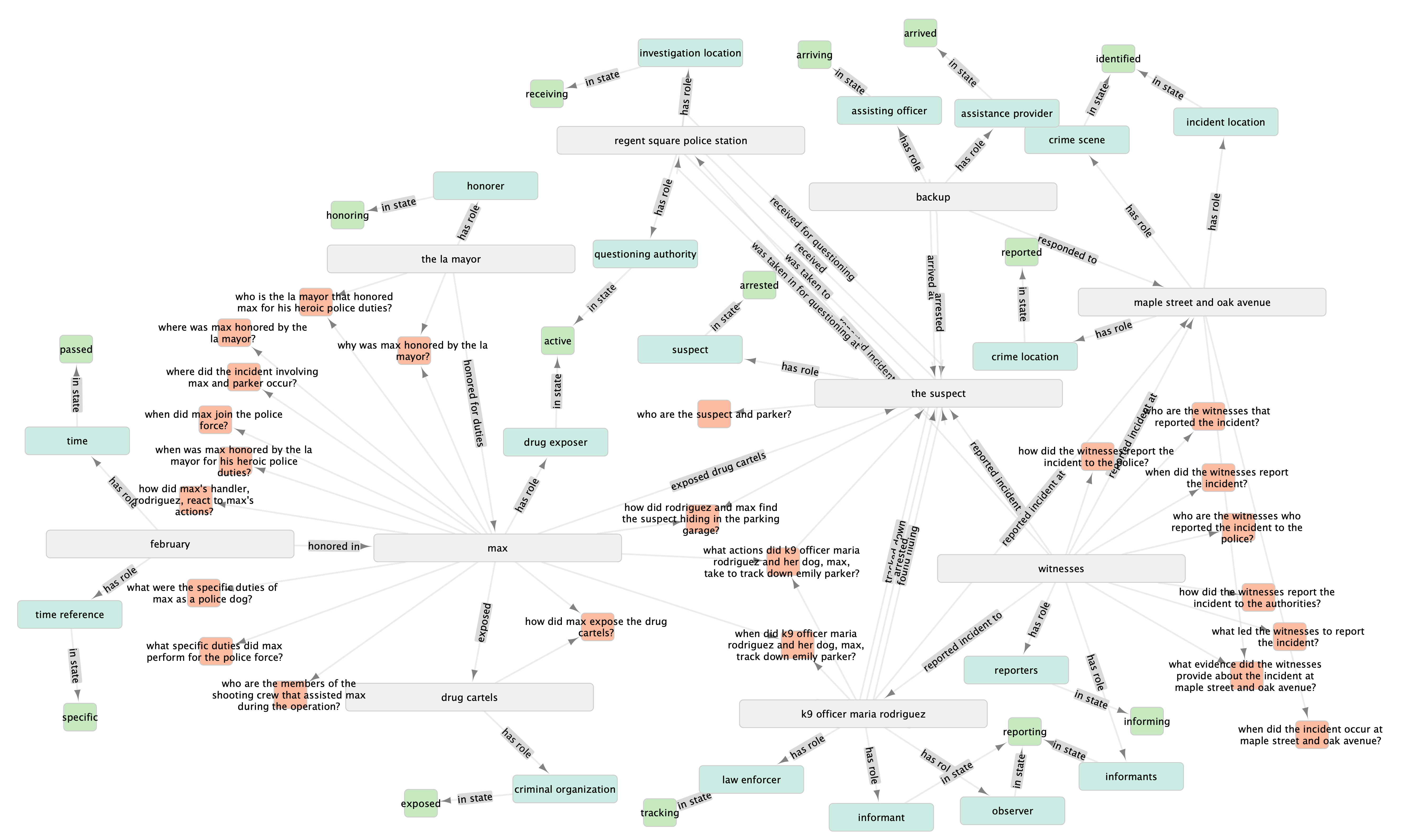}

    \caption{\textbf{Crime and Justice -} \textit{In a city hit-and-run case, K9 officer Maria Rodriguez and her dog, Max, tracked down Emily Parker, 28, who fled the scene. Witnesses reported the incident at Maple Street and Oak Avenue. Rodriguez and Max found the suspect hiding in a parking garage. Backup arrived, and Parker was arrested and taken in for questioning at Regent Square police station. Max was honored by the LA mayor for his heroic police duties in exposing drug cartels in February.}}
    \label{fig:cj2}
\end{figure*}

\begin{figure*}[ht]
    \centering
\includegraphics[trim={8cm 0 8cm 0},clip,width=\textwidth]{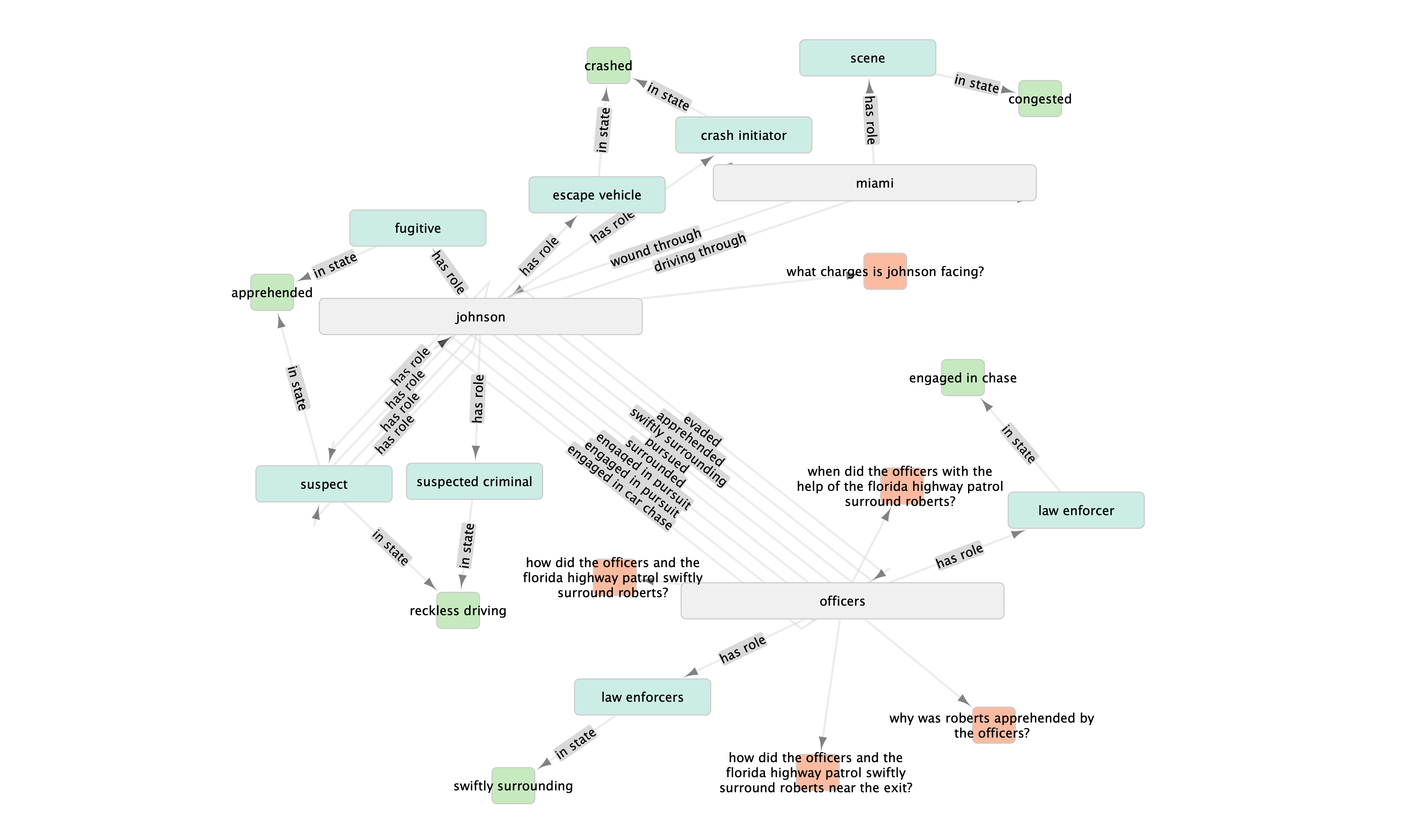}

    \caption{\textbf{Crime and Justice -} \textit{During rush hour on Interstate 95 in Miami, officers from the 23rd Precinct engaged in a harrowing car chase with Jason Roberts, 35, suspected of reckless driving and evading law enforcement. The pursuit wound through downtown Miami's crowded streets, with Johnson in a white Benz. Eventually, Roberts's vehicle careened off the interstate near exit 7B, where officers with the help of the Florida Highway Patrol swiftly surrounded him. Roberts was apprehended and taken to the Apalachian court for processing, facing charges of reckless endangerment and fleeing the scene.}}
    \label{fig:cj3}
\end{figure*}

\begin{figure*}[ht]
    \centering
\includegraphics[trim={8cm 0 8cm 0},clip,width=\textwidth]{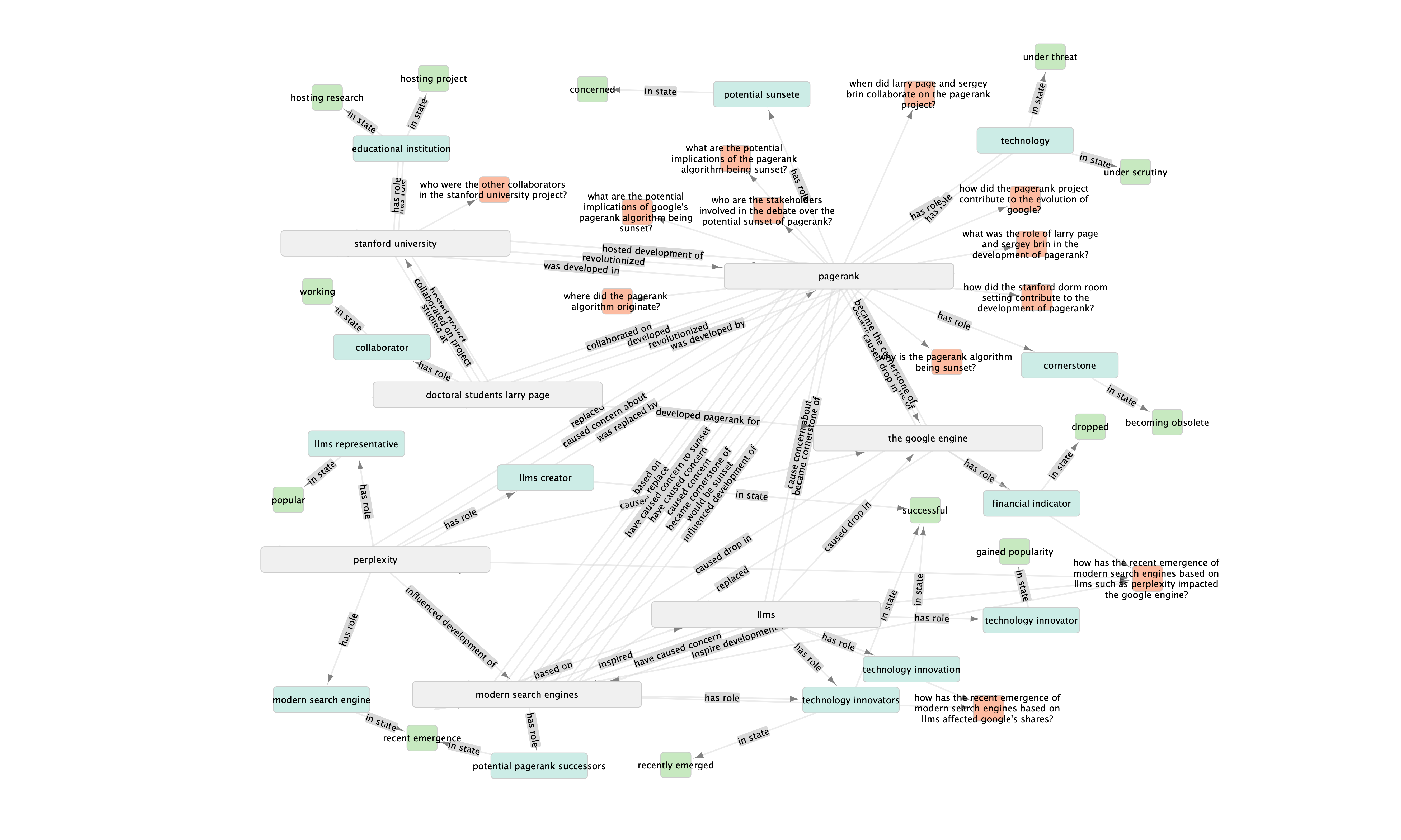}

    \caption{\textbf{Technology Development -} \textit{In the late 1990s at Stanford University, doctoral students Larry Page and Sergey Brin collaborated on a groundbreaking project called PageRank to revolutionize internet search. Developed in a Stanford dorm room, PageRank became the cornerstone of the Google engine. While initially successful, the recent emergence of modern search engines based on LLMs such as Perplexity, have caused concern that the PageRank algorithm would be sunset. Google shares have dropped 5\% since LLMs gained popularity in early 2020.}}
    \label{fig:td0}
\end{figure*}

Consider as an example
 the sentence in the caption of Fig.~\ref{fig:cj1}: 
\begin{center}
\textit{Miller allegedly brandished a weapon at a convenience store on Elm Street, demanding cash before
fleeing the scene}
\end{center} 
 the location \textit{"elm street"} is instantiated as a node with roles \textit{escape route} and \textit{crime scene}, succinctly conveying both the crime's location and the suspect fleeing the scene. Similarly, often spatial information is embedded within questions asked by the operator, in Fig.~\ref{fig:cj1} the question \textit{"what evidence did the eyewitnesses provide to the police about the incident at the convenience store?"}, conveys the fact that the location of the crime was a convenience store. Further examples including  Fig.~\ref{fig:cj2} - actor \textit{5th avenue}, Fig.~\ref{fig:cj3} - actor \textit{maple street and oak avenue}, Fig.~\ref{fig:cj3} - actor \textit{miami}, Fig.~\ref{fig:td0} - \textit{stanford university} are presented in Sec.~\ref{app:sec:examples}.
 



%

\begin{figure*}[ht]
    \centering
\includegraphics[trim={20cm 0 20cm 0},clip,width=\textwidth]{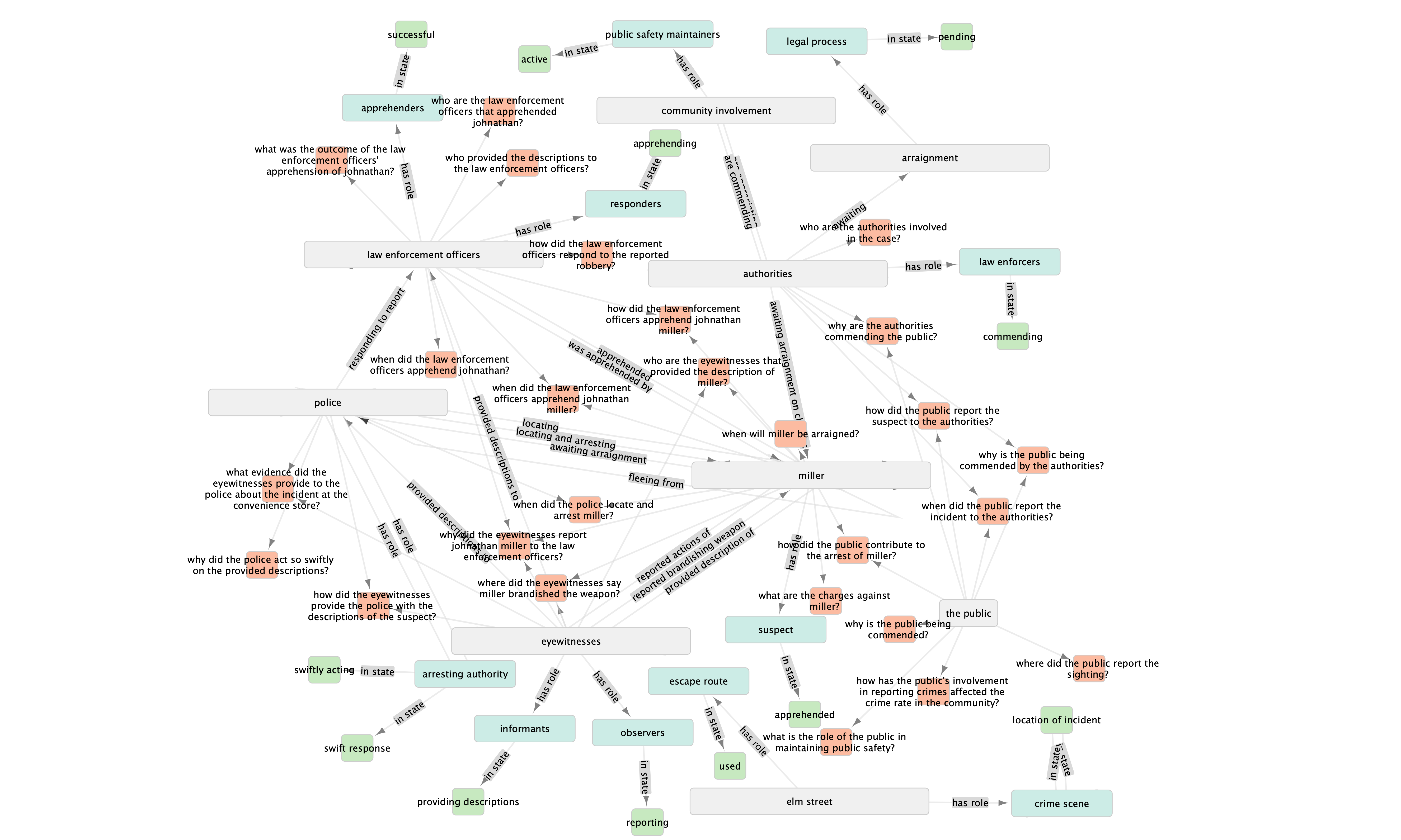}

    \caption{\textbf{Crime and Justice -} \textit{Yesterday, in a swift response to a reported robbery, law enforcement officers apprehended Johnathan Miller, a 32-year-old resident of Greenview Avenue, in the downtown area. According to eyewitnesses, Miller allegedly brandished a weapon at a convenience store on Elm Street, demanding cash before fleeing the scene. Police swiftly acted on the provided descriptions, locating and arresting Miller within the hour. He is currently detained at the city's central precinct, awaiting arraignment on charges of armed robbery. Authorities are commending the public for their quick reporting, emphasizing the importance of community involvement in maintaining public safety.}}
    \label{fig:cj0}
\end{figure*}

\begin{figure*}[ht]
    \centering
\includegraphics[trim={20cm 0 20cm 0},clip,width=\textwidth]{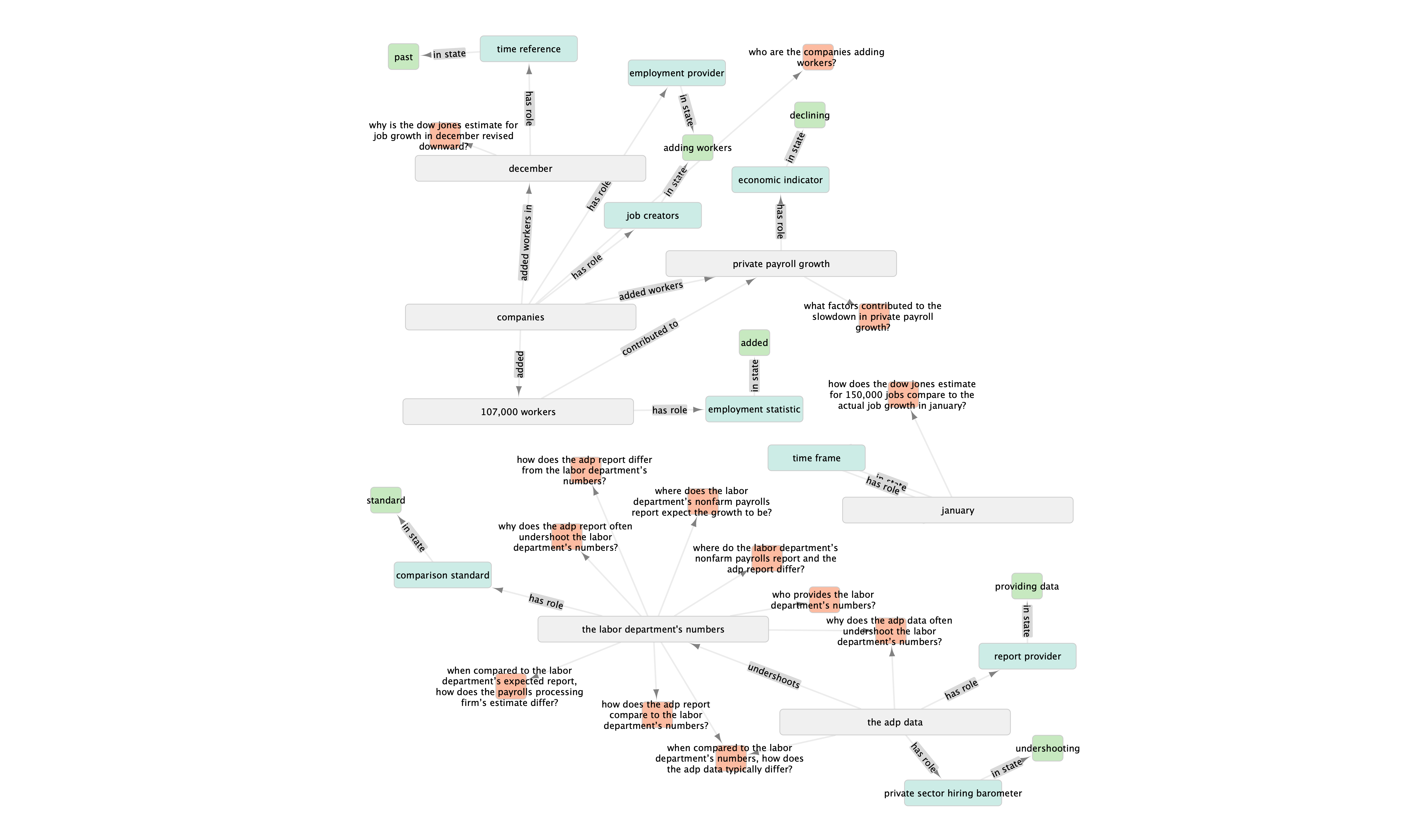}

    \caption{\textbf{Economy -} \textit{Private payroll growth declined sharply in January, a possible sign that the U.S. labor market is heading for a slowdown this year. Companies added 107,000 workers in the first month of 2024, off from the downwardly revised 158,000 in December and below the Dow Jones estimate for 150,000, according to the payrolls processing firm. The release comes two days ahead of the Labor Department’s nonfarm payrolls report, which is expected to show growth of 185,000, against the 216,000 increase in December. While the ADP data can provide a barometer for private sector hiring, the two reports often differ, with ADP often undershooting the Labor Department’s numbers.}}
    \label{fig:eco0}
\end{figure*}

\begin{figure*}[ht]
    \centering
\includegraphics[trim={10cm 0 10cm 0},clip,width=\textwidth]{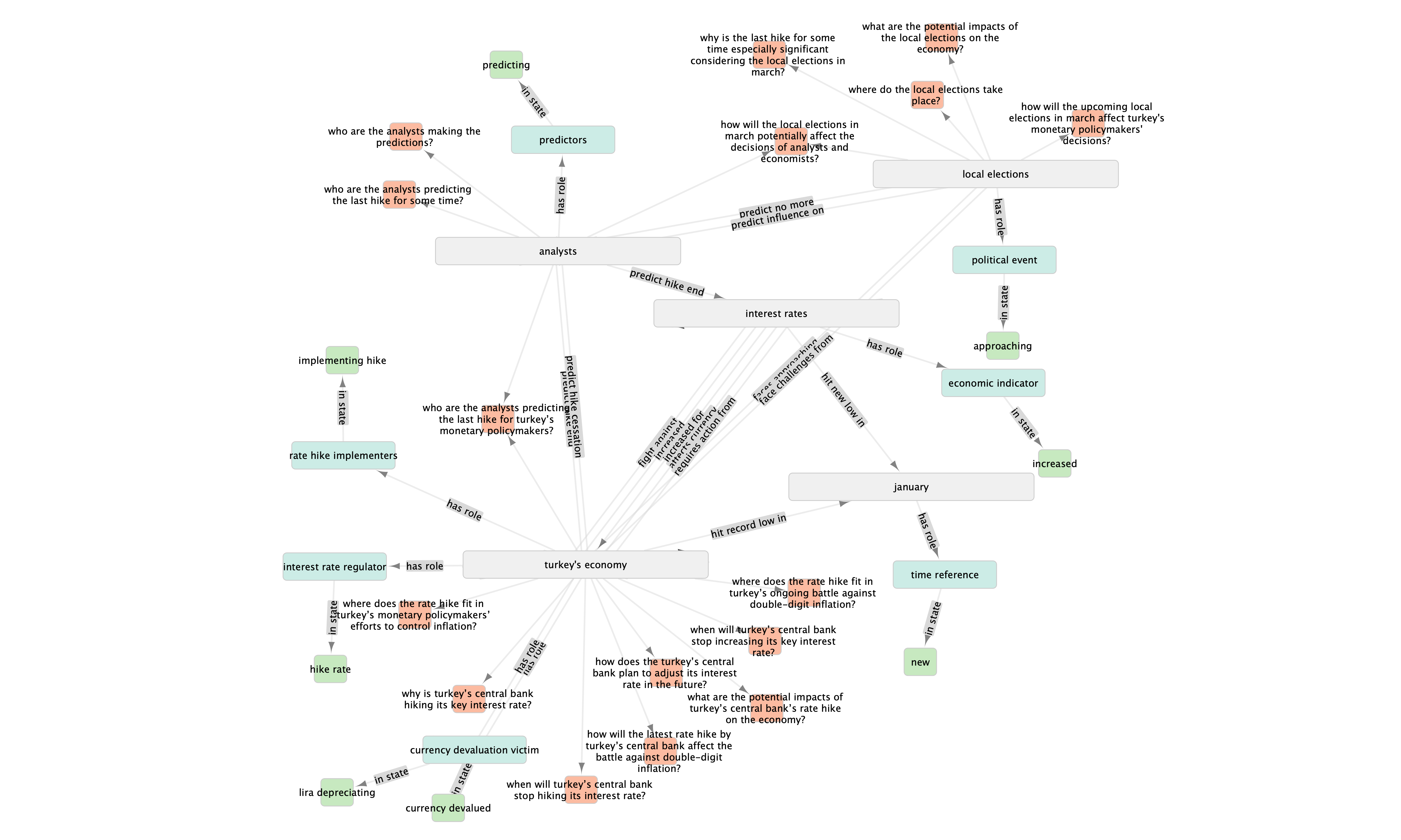}

    \caption{\textbf{Economy -} \textit{Turkey’s central bank on Thursday hiked its key interest rate by another 250 basis points to 45\%. The hike to the benchmark one-week repo rate was in line with economists’ expectations. It comes amid an ongoing battle against double-digit inflation for Turkey’s monetary policymakers, with the rate hike the latest step in that effort. Inflation in Turkey increased to 64.8\% year-on-year in December, up from 62\% in November, and the country’s currency, the lira, hit a new record low against the U.S. dollar earlier in January, breaking 30 to the greenback for the first time. Analysts predict this will be the last hike for some time, especially with local elections approaching in March.}}
    \label{fig:eco3}
\end{figure*}

\noindent \textbf{Capturing Temporal Information:} As with spatial information, the operator represents explicit temporal details as actors with associated roles and states. Furthermore, implicit temporal information is often encoded in the roles and states of other actors, suggesting the progression or completion of an action. For example, in the caption for Fig.~\ref{fig:eco3}: 

\begin{center}
   \textit{the lira, hit a new record low against the U.S. dollar earlier in January, breaking 30 to the
greenback for the first time. Analysts predict this will be the last hike for some time, especially with local elections
approaching in March.}
\end{center}

text spans like \textit{"January"}, indicating the time frame of occurance of an event, are captured as Nodes in the network, and, more implicit temporal information, is conveyed through the States accompanying the action - in this case state \textit{approaching} of actor \textit{local elections}.  Further examples including  Fig.~\ref{fig:cj0} - state \textit{pending} of actor \textit{arraignemnt}, Fig.~\ref{fig:cj3} - actor \textit{february}, Fig.~\ref{fig:eco0} - actor \textit{december} are presented in Sec.~\ref{app:sec:examples}.




At the Reconciler-level, temporal and spatial reasoning is paramount to deciding whether nodes and edges from the latest Operator workspace instance $\mathcal{W}_n$ should replace, augment to, or be rejected in favor of the instance in the Working memory $\mathcal{M}_{n}^*$. Future work will attempt to highlight these time and space features embedded in the Workspace instances and project the various events and inter-actor interactions described in the Workspace onto a shared timeline and virtual space.

\section{Convergence Plots}

\begin{figure}[ht]
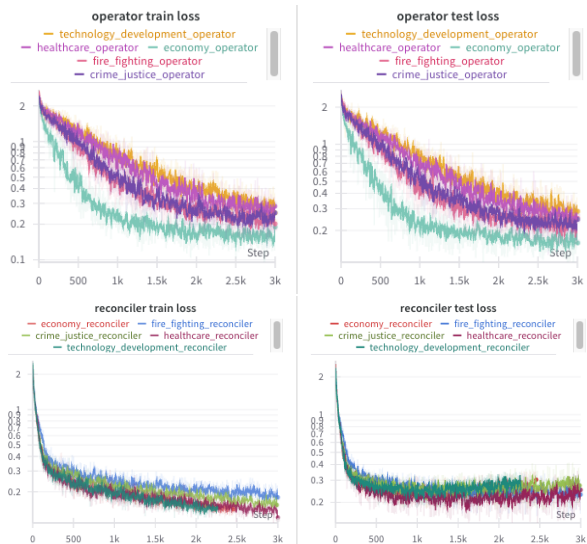

    \centering
    \includegraphics[width=\columnwidth]{Figures/operator\_convergence.png}
    \includegraphics[width=\columnwidth]{Figures/reconciler\_convergence.png}
    \caption{\textbf{Operator and Reconciler convergence plots across situations}: Each operator/reconciler model uses shared LLaMA2 (backbone) + QLoRA weights -- \textit{plug-and-play setup}. Strong train and test loss convergence implies task learning and good generalizing potential.}
    \label{fig:convergences}
\end{figure}

Convergence loss curves for all situations are reported in Fig.~\ref{fig:convergences}.

\section{Annotator Guidelines} \label{app:annotator}

\begin{figure*}[ht!]
\begin{minipage}{1.0\textwidth}
\centering
\includegraphics[width=\textwidth]{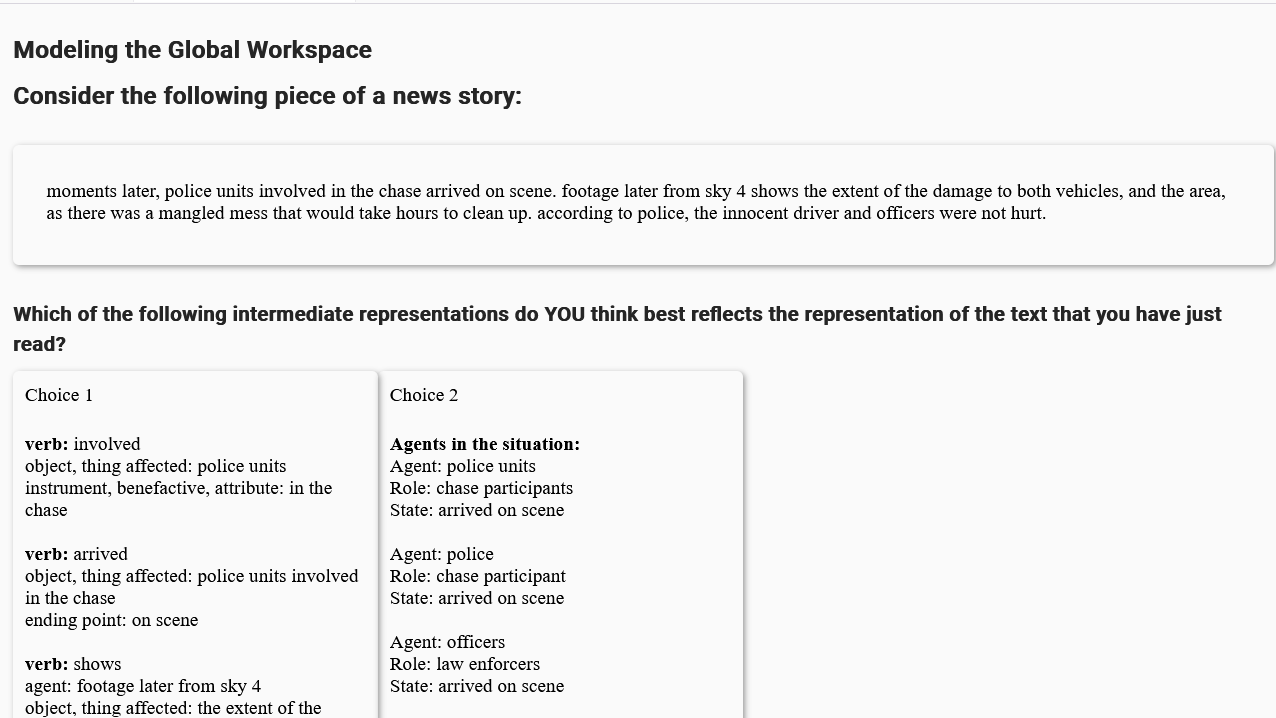}
\caption{\textbf{Annotator instructions for UpWork Task 1:} Annotators are asked to compare the outputs of the Operator to the Semantic map output by a baseline framework (either GLEN, BertSRL, FST) given a shared input text context. During annotation, one random baseline map and the Operator output are presented in random order and the annotator is asked to pick the representation of the Semantics that best reflects the information in the context.}
\label{fig:ann1}
\end{minipage}

\vspace{2em}

\begin{minipage}{1.0\textwidth}
\centering
\includegraphics[width=\textwidth]{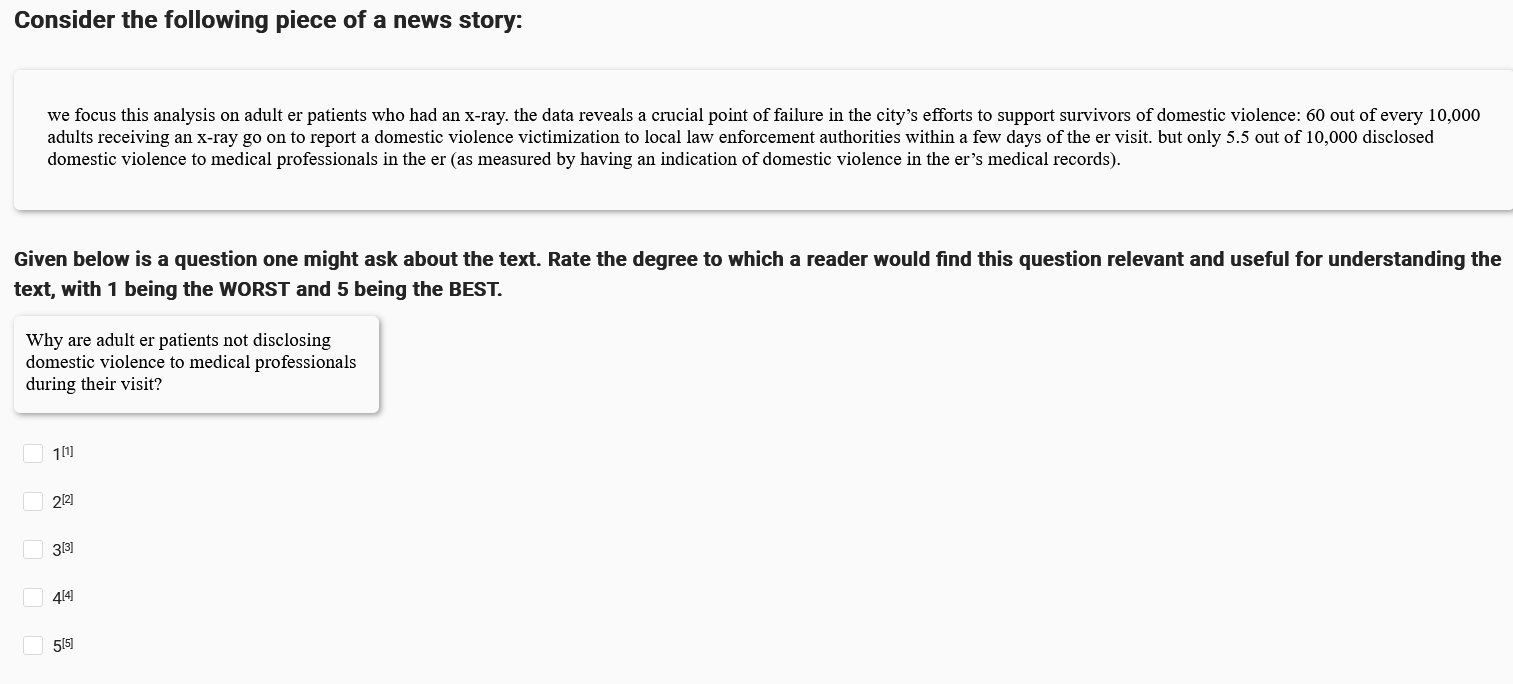}
\caption{\textbf{Annotator instructions for UpWork Task 2:} Given a text segment, annotators are faced with the task of comparing the generated questions from GSW (Operator model) to the questions generated from GPT-4 (annotators are unaware during annotation which question is generated by the GSW model). Annotators find it extremely difficult to identify the source model.}
\label{fig:ann2}
\end{minipage}
\end{figure*}

Annotators who exceeded $\$50K$ in total gross pay were recruited from UpWork, a talent resource. These candidates were first interviewed in a $10$-minute session to verify that they were proficient in English and those that had prior experience in annotating large-scale AI/ML data -- listed as a verified skill on the platform -- were selected to move on to the next round. Following this, they were given a set of $10$ task prototype examples and $10$ unanswered labeling tasks. Those that got $9$ out of the $10$ annotations right moved on to the first round of labeling. Each task was labeled twice -- by the \textit{annotator} and a \textit{verifier} -- to ensure quality of the results. Annotators were paid $\$5 / 40 \textrm{ samples}$ which was estimated to take them about $30$ minutes at most, or at the rate of $\$10 / \textrm{hour}$, which was confirmed to exceed the federal minimum wage where the annotators were situated. The raw annotations are to be released as part of the codebase. Raw annotator guidelines are presented in Figs.~\ref{fig:ann1},\ref{fig:ann2}.

\section{Ablation} \label{sec:ablation}

\begin{table}[ht!]
\centering
\small
\setlength\extrarowheight{-4pt}
\begingroup
\renewcommand{\arraystretch}{0.6} 
\begin{tabular}{@{}cccc@{}}
\toprule
\multicolumn{1}{c|}{\textbf{Base Model}} &
  \multicolumn{1}{c|}{\textbf{Adapters}} &
  \multicolumn{1}{c|}{\textbf{Precision}} &
  \textbf{Parameters} \\ \midrule

\multicolumn{1}{c|}{LLaMA-7b}            & \multicolumn{1}{c|}{LoRA}  & \multicolumn{1}{c|}{16 bit} & 1M   \\ \midrule
\multicolumn{1}{c|}{\textbf{Mistral-7b}} & \multicolumn{1}{c|}{QLoRA} & \multicolumn{1}{c|}{4 bit}  & 800k \\ \midrule
\rowcolor[HTML]{D9F7D1} 
\multicolumn{1}{c|}{\cellcolor[HTML]{D9F7D1}\textbf{LLaMA-13b}} &
  \multicolumn{1}{c|}{\cellcolor[HTML]{D9F7D1}\textbf{QLoRA}} &
  \multicolumn{1}{c|}{\cellcolor[HTML]{D9F7D1}\textbf{4 bit}} &
  \textbf{1.6M} \\ \midrule

\multicolumn{1}{c|}{Llama-XXb Chat}       & \multicolumn{1}{c|}{-}     & \multicolumn{1}{c|}{-}      & -    \\ \bottomrule
\end{tabular}
\endgroup
\caption{\textbf{Model Architecture Search:} LLaMA-13b trained with 4-bit QLoRA was found to output significantly better \textit{Crime and Justice} situation-specific workspace instances. Finetuned T5~\cite{T5}, Mistral~\cite{mistral} (7B) models performed worse. LLaMA 70B models performed comparably to the 13B parameter model. LLaMA-Chat models (7B, 13B, 70B) demonstrate poor quality outputs with no further finetuning. }
\label{tab:ablation-model-arch}
\end{table}

\begin{table*}[t!]
\centering
\small\addtolength{\tabcolsep}{+5.2pt}
\begingroup
\renewcommand{\arraystretch}{0.6} 
\begin{tabular}{@{}cc|ccc|cc@{}}
\toprule
\multicolumn{2}{c|}{\textbf{Situation}}                                                            & \multicolumn{3}{c|}{\textbf{Reconciliation}}                                   & \multicolumn{2}{c}{\textbf{Question Resolution}}   \\ \midrule
\multicolumn{1}{c|}{Train}                                       & Test                            & \multicolumn{1}{c|}{ACC.$\uparrow$} & \multicolumn{1}{c|}{F1$\uparrow$}            & Sensitivity$\uparrow$   & \multicolumn{1}{c|}{ACC.$\uparrow$}          & F1$\uparrow$            \\ \midrule
\multicolumn{1}{c|}{\multirow{5}{*}{\textit{crime and justice}}} & \textit{crime and justice}      & \multicolumn{1}{c|}{0.81} & \multicolumn{1}{c|}{0.81}          & 0.91          & \multicolumn{1}{c|}{0.92}          & 0.92          \\ \cmidrule(l){2-7} 
\multicolumn{1}{c|}{}                                            & \textit{fire fighting}          & \multicolumn{1}{c|}{0.83} & \multicolumn{1}{c|}{0.83}          & 0.95          & \multicolumn{1}{c|}{0.95}          & 0.94          \\ \cmidrule(l){2-7} 
\multicolumn{1}{c|}{}                                            & \textit{technology development} & \multicolumn{1}{c|}{0.87} & \multicolumn{1}{c|}{\textbf{0.87}} & \textbf{0.94} & \multicolumn{1}{c|}{0.99}          & 0.99          \\ \cmidrule(l){2-7} 
\multicolumn{1}{c|}{}                                            & \textit{healthcare}             & \multicolumn{1}{c|}{0.85} & \multicolumn{1}{c|}{0.85}          & 0.92          & \multicolumn{1}{c|}{0.92}          & 0.92          \\ \cmidrule(l){2-7} 
\multicolumn{1}{c|}{}                                            & \textit{economy}                & \multicolumn{1}{c|}{0.80} & \multicolumn{1}{c|}{0.81}          & 0.87          & \multicolumn{1}{c|}{\textbf{0.99}} & \textbf{0.99} \\ \bottomrule
\end{tabular}
\endgroup
\caption{\textbf{Generalization of one Reconciler to other Workspaces}: A Reconciler trained on the \textit{Crime and Justice} situation is shown to generalize across the other situations. Bold indicates cases where classification performance of the \textit{Crime and Justice} trained reconciler is better than the performance of reconcilers trained on each specific situation (compare to Tab.~\ref{tab:all-reconciler}).}
\label{tab:ablation-table}
\end{table*}

\noindent \textit{A.1 Generalizing Operator and Reconciler models to other situations:} The \textit{Crime and Justice} Operator and Reconciler models are evaluated on text segments associated to \textit{other situations}. We observe very little ``situation-bias'' in the lexicon (specifically the roles, states and questions) generated by the Operator -- example shown in App. Sec.~\ref{fig:cross}. Additionally, the \textit{Crime and Justice} Reconciler shows strong classification performance \textit{across all situations}, and even outperforms the situation-specific models in some cases (see Tab.~\ref{tab:ablation-table}, in bold). 


\noindent \textit{A.2 Can LLaMA-based Chat models perform the role of the Operator and the Reconciler without any custom training?} In Tab.~\ref{tab:ablation-model-arch}, we show the alternate model architectures and PeFT adapters considered for training the Operator and Reconciler models. We manually evaluated the quality of the outputs of the trained models and the prompted Chat versions, and these performed significantly worse than the architecture we have used in the main text (highlighted in green).

\section{Situation transfer from one situation to another} \label{fig:cross}

The following Workspace instance is generated by the Crime and Justice-trained LoRA model when applied on the following post from a \textit{Fire-Fighting} situation:

\begin{center}
    \textit{
    After a gender reveal party gone wrong, 10 units of the LAFD were sent to handle the fire at the scene. The gender reveal party caused over a million acres to the engulfed in flames in South Savannah, causing billions of dollars in damages and loss of property.}
\end{center}

\newpage

\noindent \textbf{Nodes(actor, role, states):} \\
"actor": "10 units", "role": "firefighting force", "state": "deployed", \\
"actor": "flames", "role": "cause of destruction", "state": "engulfing", \\
"actor": "party", "role": "damage causer", "state": "caused damage", \\
"actor": "property", "role": "loss victim", "state": "damaged", \\
"actor": "south savannah", "role": "affected area", "state": "engulfed", \\

\noindent \textbf{Edges(label, source, target):} \\
"label": "caused loss of", "source": "flames", "target": "property", "attributes": "in south savannah" \\
"label": "engulfed in", "source": "flames", "target": "south savannah", "attributes": "in south savannah" \\
"label": "caused loss of", "source": "party", "target": "property", "attributes": "in south savannah" \\
"label": "caused fire in", "source": "party", "target": "south savannah", "attributes": "south savannah" \\

\noindent \textbf{Questions:}\\
"who was responsible for causing the fire at the gender reveal party in south savannah?", \\
"how did the gender reveal party lead to the destruction of over a million acres in south savannah?", \\
"when and where did the gender reveal party take place?", "where did the 10 units of the lafd go after responding to the fire?", \\
"what measures are being taken to prevent such incidents in the future?", \\
"why did the 10 units of the lafd respond to the fire at the scene?"
\color{black}

\section{Sample Workspace Instances} \label{app:sec:examples}

In the next few pages, we present $4$ workspace instances for each situation in Tab.~\ref{tab:data} generated by the Workspace (Operator+Reconciler) model at the end of a given passage (mentioned in the caption). These figures can be considered as additional results supporting the one instance shown in Fig.~\ref{fig:spotlight}.

\newpage

\newpage

\newpage

\newpage

\begin{figure*}[ht]
    \centering
\includegraphics[trim={0cm 0 0cm 0},clip,width=\textwidth]{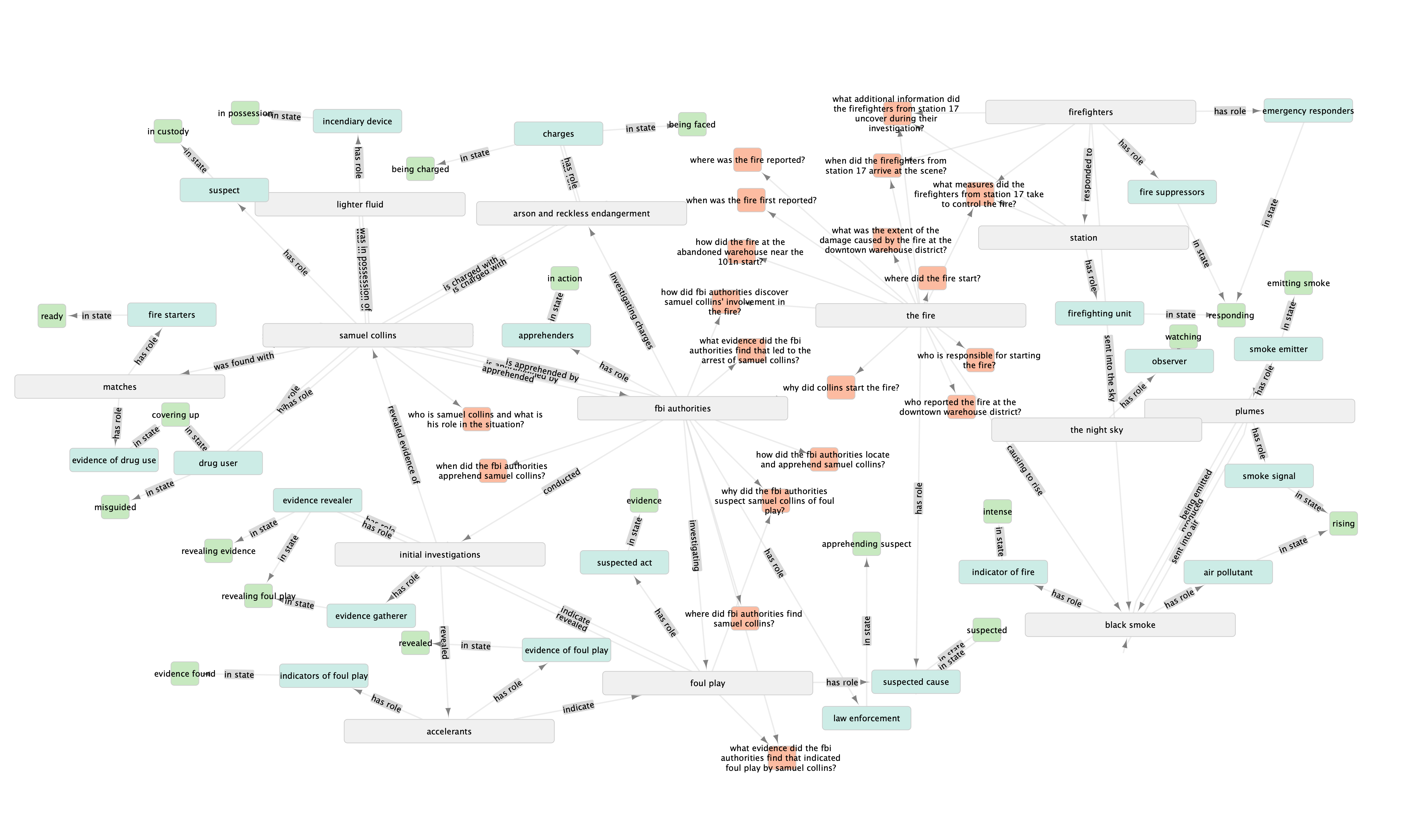}

    \caption{\textbf{Fire Fighting -} \textit{In a swift response to a blaze reported at the downtown warehouse district, firefighters from Station 17 battled a ferocious fire that erupted at an abandoned warehouse near the 101N late last night. The fire, suspected to be arson, engulfed the entire structure on 5th Street, sending plumes of black smoke into the night sky. Initial investigations have revealed evidence of accelerants, indicating foul play. FBI authorities apprehended Samuel Collins, 24, discovered under a moving van parked nearby, with matches and lighter fluid in his possession. Collins is believed to have started the fire in a misguided attempt to cover up evidence of drug use, and faces charges of arson and reckless endangerment. The fire is 90\% contained.}}
    \label{fig:ff0}
\end{figure*}

\newpage

\begin{figure*}[ht]
    \centering
\includegraphics[trim={0cm 0 0cm 0},clip,width=\textwidth]{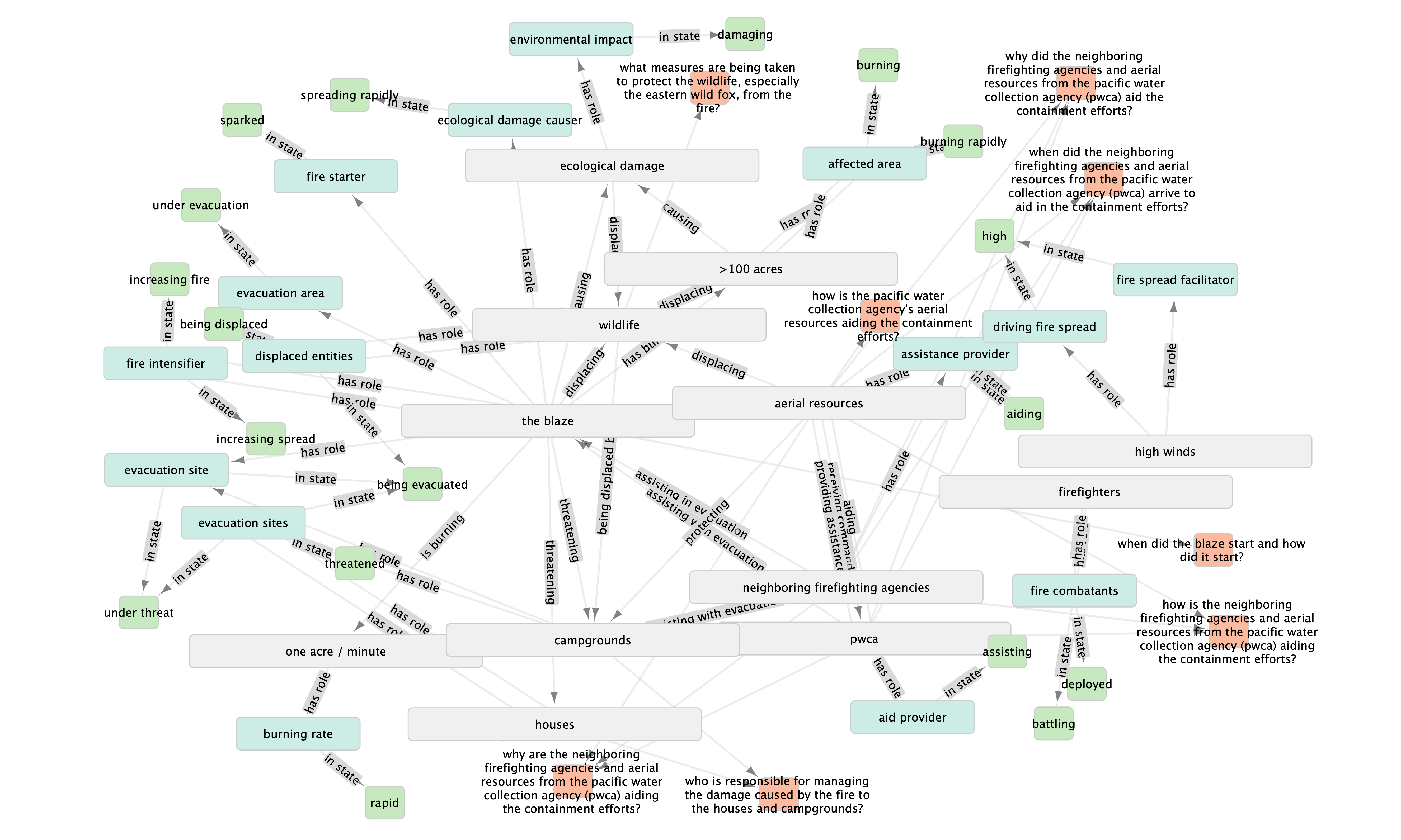}

    \caption{\textbf{Fire Fighting -} \textit{Firefighters from the Redwood County Fire Department are battling a forest fire sparked by lightning in the eastern part of the Redwood National Forest. The blaze is spreading rapidly (estimated to burn through one acre / minute) due to high winds and dry conditions. Evacuations are underway at the nearby Cahuenga community, as the fire threatens houses and campgrounds. Despite efforts to contain it, the fire continues to spread (having burned >100 acres), causing ecological damage and displacing wildlife, especially the Eastern wild fox. Assistance from neighboring firefighting agencies and aerial resources from the Pacific Water Collection Agency (PWCA) is aiding containment efforts.}}
    \label{fig:ff1}
\end{figure*}

\newpage

\begin{figure*}[ht]
    \centering
\includegraphics[trim={0cm 0 120cm 20cm},clip,width=\textwidth]{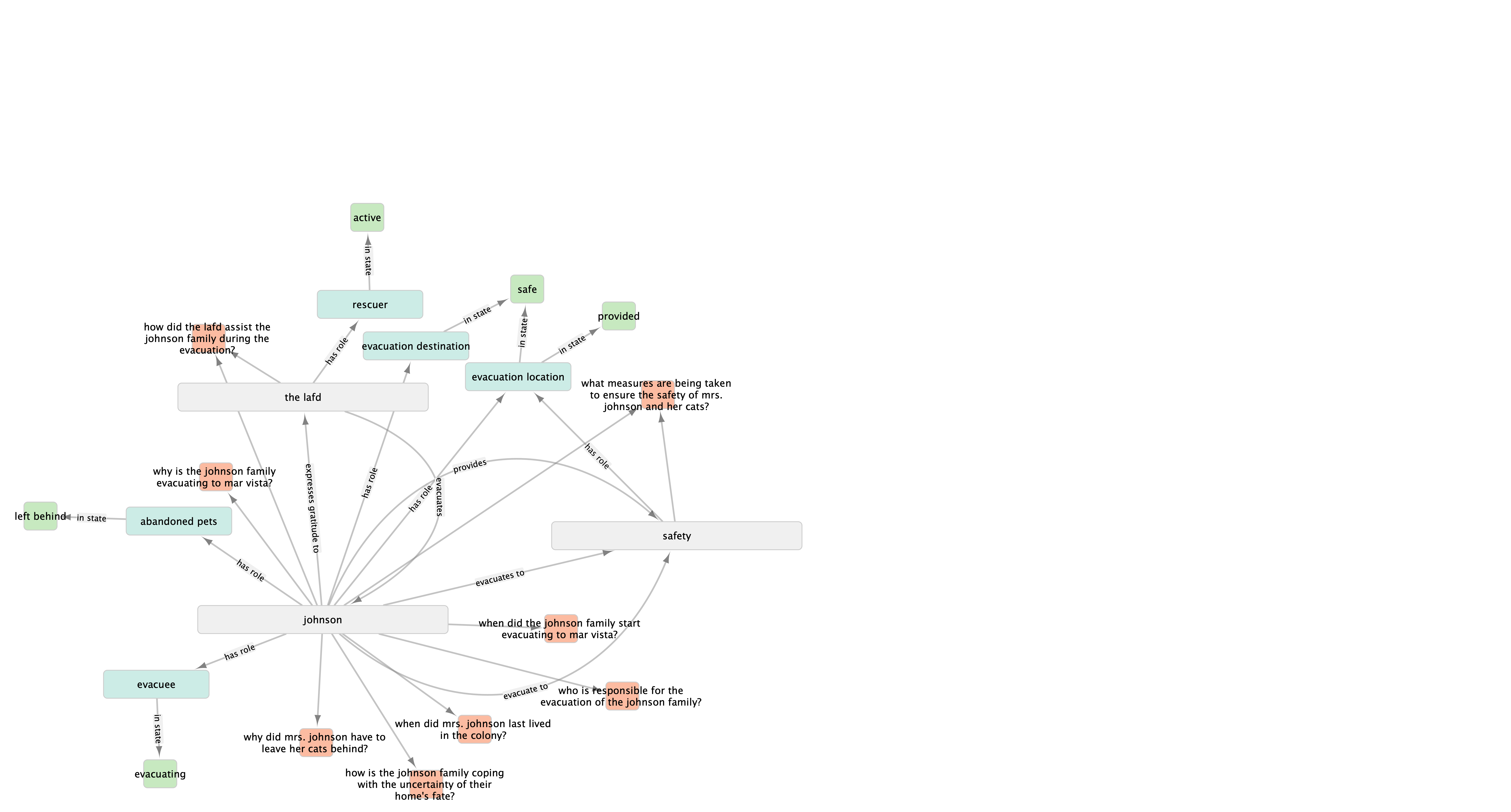}

    \caption{\textbf{Fire Fighting -} \textit{In the Redwood National Forest, an electric fire sparked a larger wildfire that now has spread to over 50 acres and threatens over a hundred homes and indigeneous wildlife. Many families including the Johnson family anxiously awaits news of their home's fate as they evacuate to safety to Mar Vista. Mrs. Johnson was visibly distraught: 'I am grateful for the LAFD for evacuating me. I hope the colony remains intact. I have lived there for over 60 years and I had to leave my 3 cats behind.' Respite may take a while however with dry conditions expected from the El Nino weather patterns over the next two weeks.}}
    \label{fig:ff2}
\end{figure*}

\newpage

\begin{figure*}[ht]
    \centering
\includegraphics[trim={5cm 0 5cm 0},clip,width=\textwidth]{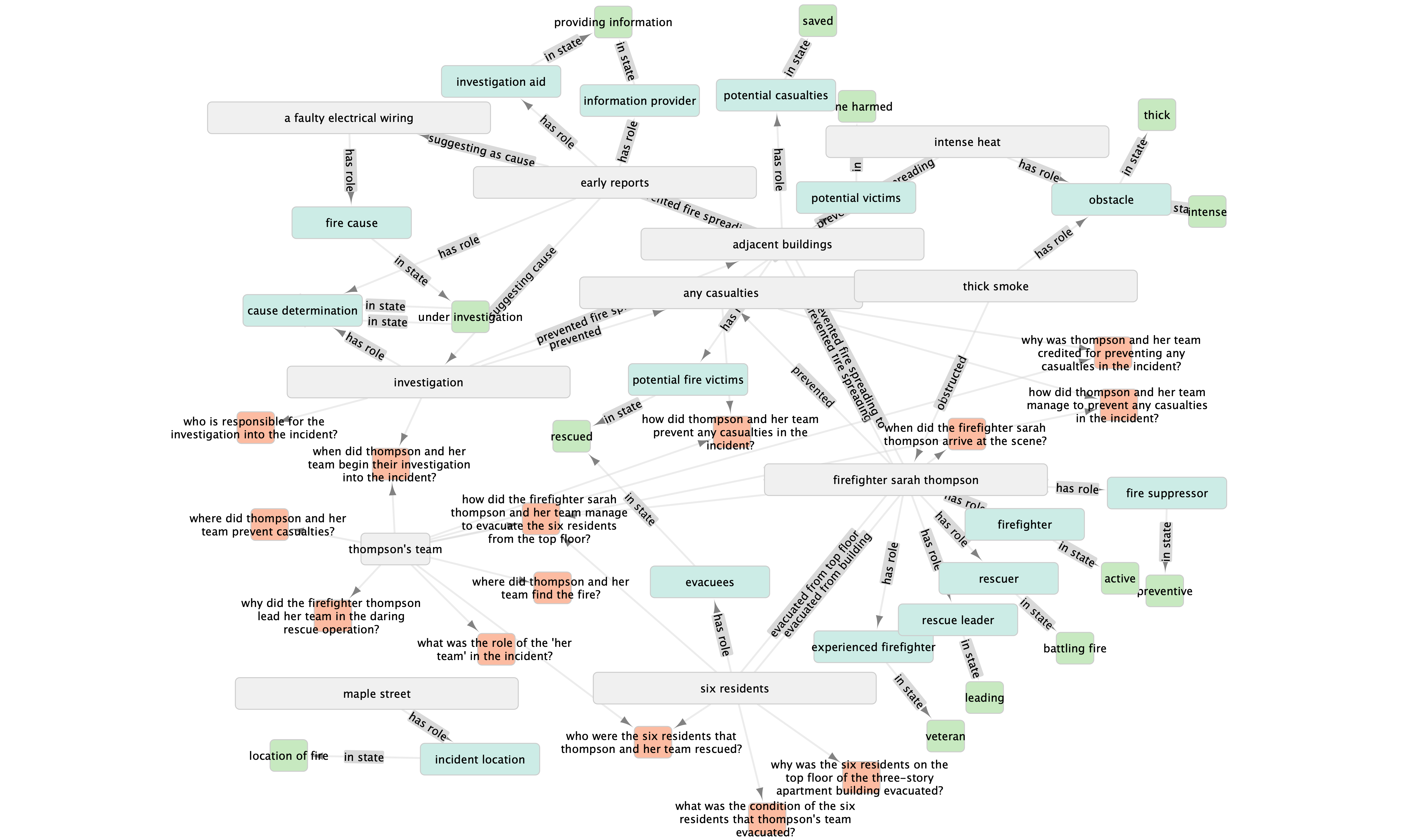}

    \caption{\textbf{Fire Fighting -} \textit{In downtown Greenville, firefighter Sarah Thompson battled a fierce blaze that engulfed a three-story apartment building on Maple Street late last night. Thompson, a veteran of the Greenville Fire Department, led her team in a daring rescue operation, evacuating six residents trapped on the top floor. Despite intense heat and thick smoke, Thompson's swift actions prevented the fire from spreading to adjacent buildings. The cause of the fire is under investigation, with early reports suggesting a faulty electrical wiring. Thompson and her team are credited with preventing any casualties in the incident.}}
    \label{fig:ff3}
\end{figure*}

\newpage

\begin{figure*}[ht]
    \centering
\includegraphics[trim={0cm 0 0cm 0},clip,width=\textwidth]{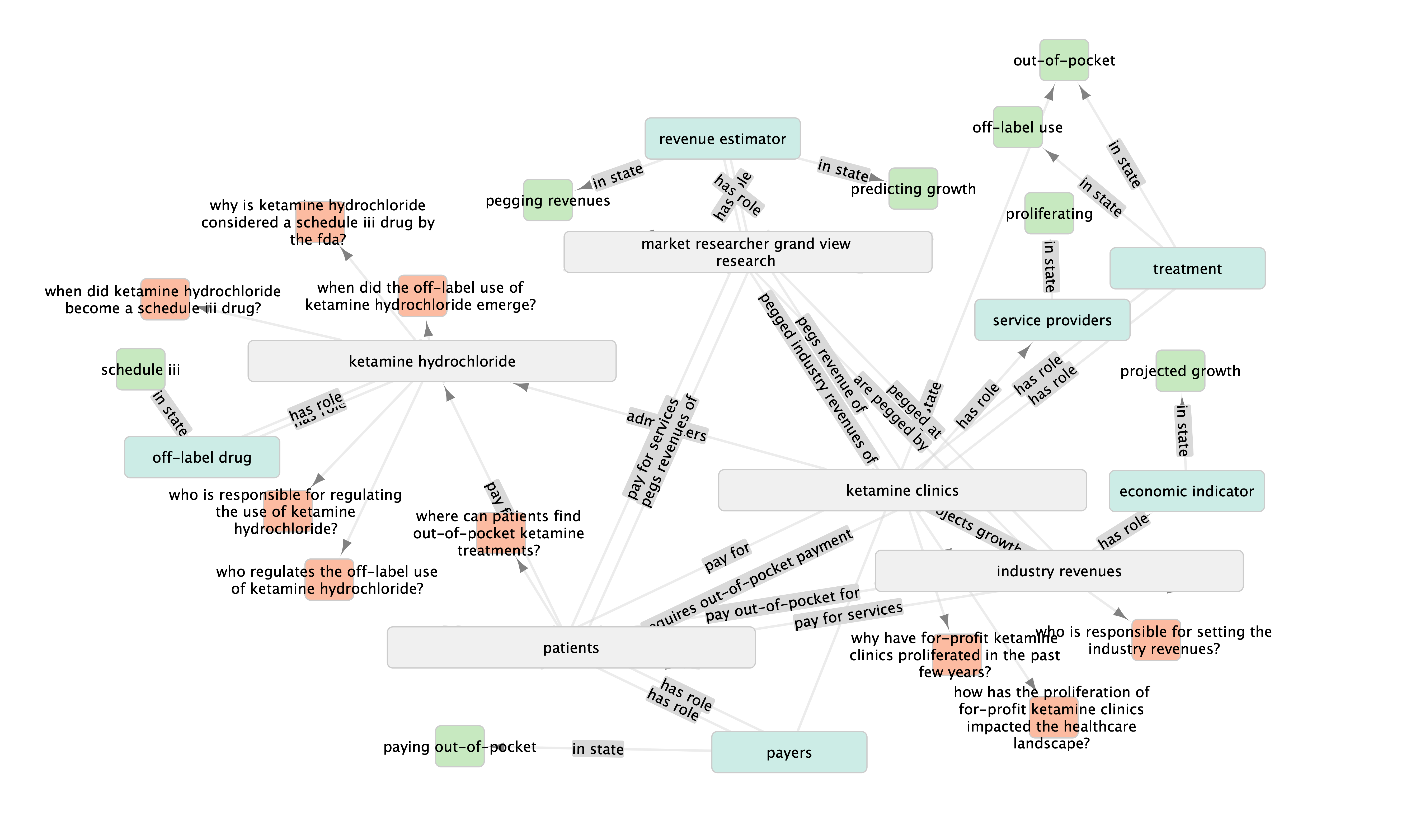}

    \caption{\textbf{Healthcare -} \textit{For-profit ketamine clinics have proliferated over the past few years, offering infusions for a wide array of mental health issues, including obsessive-compulsive disorder, depression, and anxiety. Although the off-label use of ketamine hydrochloride, a Schedule III drug approved by the FDA as an anesthetic in 1970, was considered radical just a decade ago, now between 500 and 750 ketamine clinics have cropped up across the nation. Market researcher Grand View Research pegged industry revenues at \$3.1 billion in 2022, and projects them to more than double to \$6.9 billion by 2030. Most insurance doesn’t cover ketamine for mental health, so patients must pay out-of-pocket.}}
    \label{fig:hc0}
\end{figure*}

\newpage

\begin{figure*}[ht]
    \centering
\includegraphics[trim={0cm 0 0cm 0},clip,width=\textwidth]{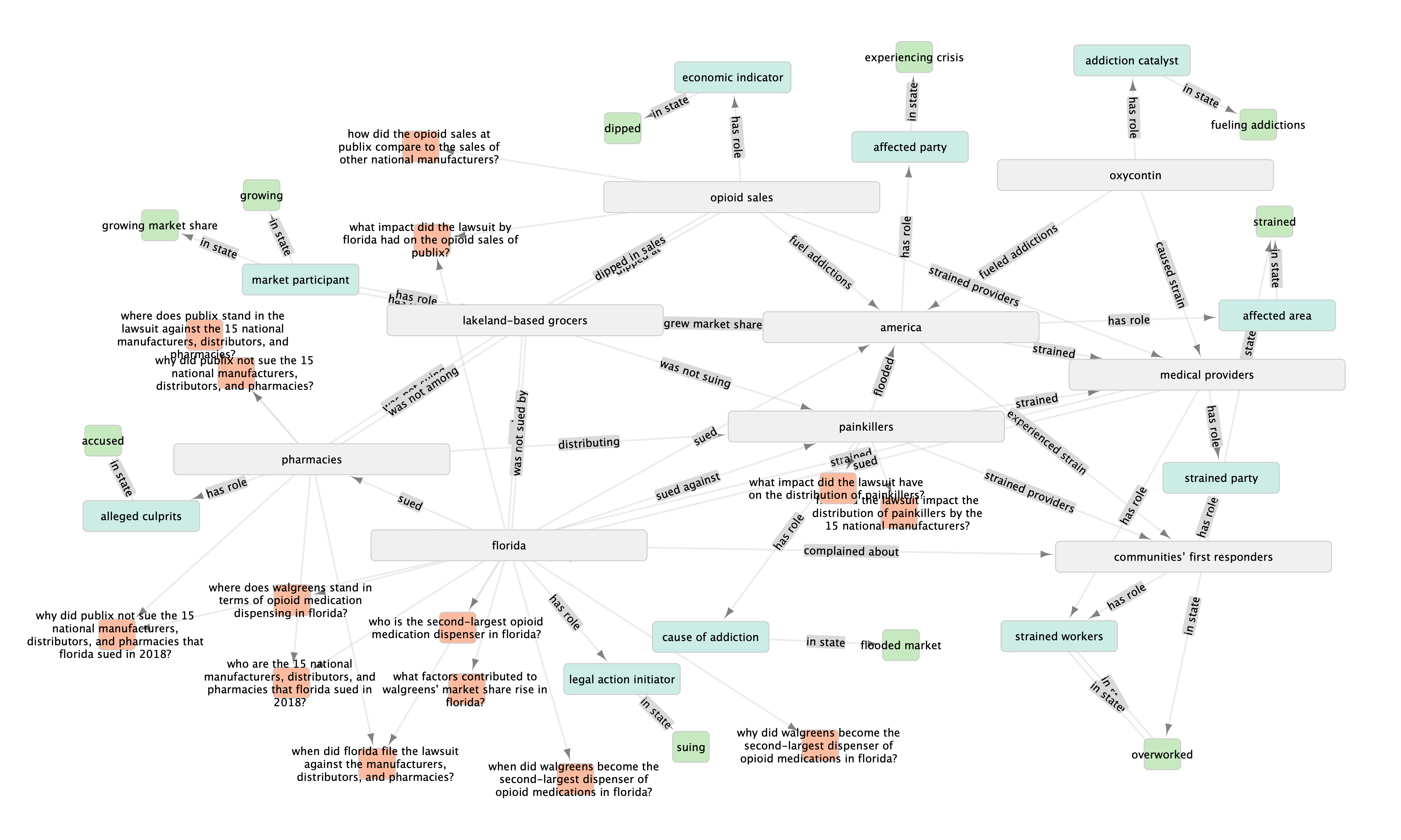}

    \caption{\textbf{Healthcare -} \textit{	"The Lakeland-based grocer’s sales of oxycodone climbed from 26 million pills per year in 2011 to 43.5 million in 2019, the data shows. The increase in sales, which far outpaced the chain’s addition of stores in Florida, saw its market share rise to 14\%, enough to overtake CVS to become Florida's second-largest dispenser of all opioid medications, behind only Walgreens, which dispensed 28\% of opioids in the state in 2019. The analysis excludes drugs like methadone prescribed for addiction treatment. Opioid sales at Publix dipped slightly in 2018 and 2019, the last two years with available data. Even as its market share grew, however, Publix was not among the 15 national manufacturers, distributors, and pharmacies that Florida sued in 2018. That lawsuit claimed other pharmacies had flooded America with painkillers such as OxyContin, fueling debilitating addictions that strained communities’ first responders and medical providers.}}
    \label{fig:hc1}
\end{figure*}

\newpage

\begin{figure*}[ht]
    \centering
\includegraphics[trim={12cm 0 12cm 0},clip,width=\textwidth]{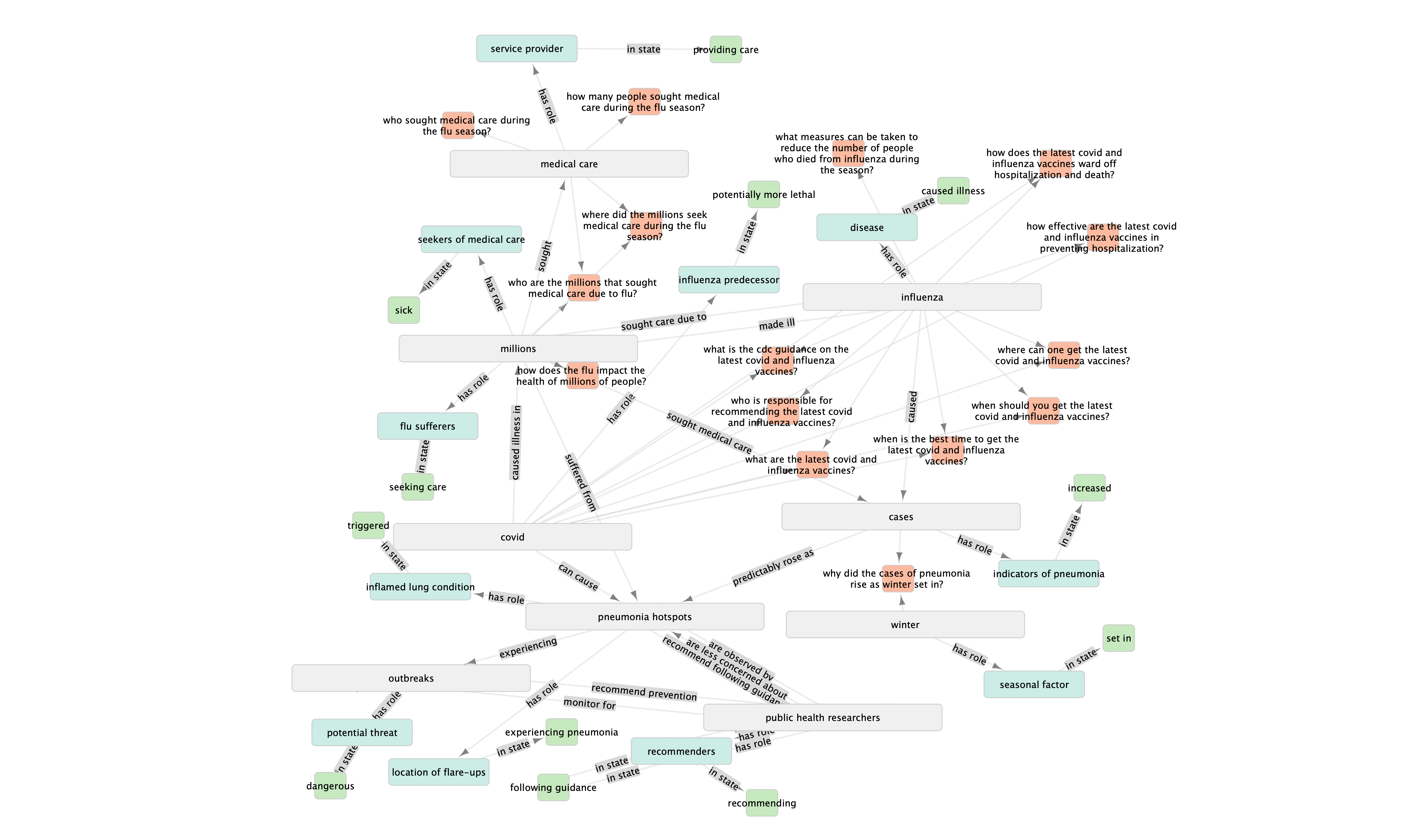}

    \caption{\textbf{Healthcare -} \textit{Between 9,400 and 28,000 people died from influenza from Oct. 1 to Jan. 6, estimates the Centers for Disease Control and Prevention, and millions felt so ill from the flu that they sought medical care. Cases of pneumonia — a serious condition marked by inflamed lungs that can be triggered by the flu, covid, or other infections — also predictably rose as winter set in. Researchers are now less concerned about flare-ups of pneumonia in China, Denmark, and France in November and December, because they fit cyclical patterns of the pneumonia-causing bacteria Mycoplasma pneumoniae rather than outbreaks of a dangerous new bug. Public health researchers recommend following the CDC guidance on getting the latest covid and influenza vaccines to ward off hospitalization and death from the diseases and reduce chances of getting sick.}}
    \label{fig:hc2}
\end{figure*}

\newpage

\begin{figure*}[ht]
    \centering
\includegraphics[trim={5cm 0 5cm 0},clip,width=\textwidth]{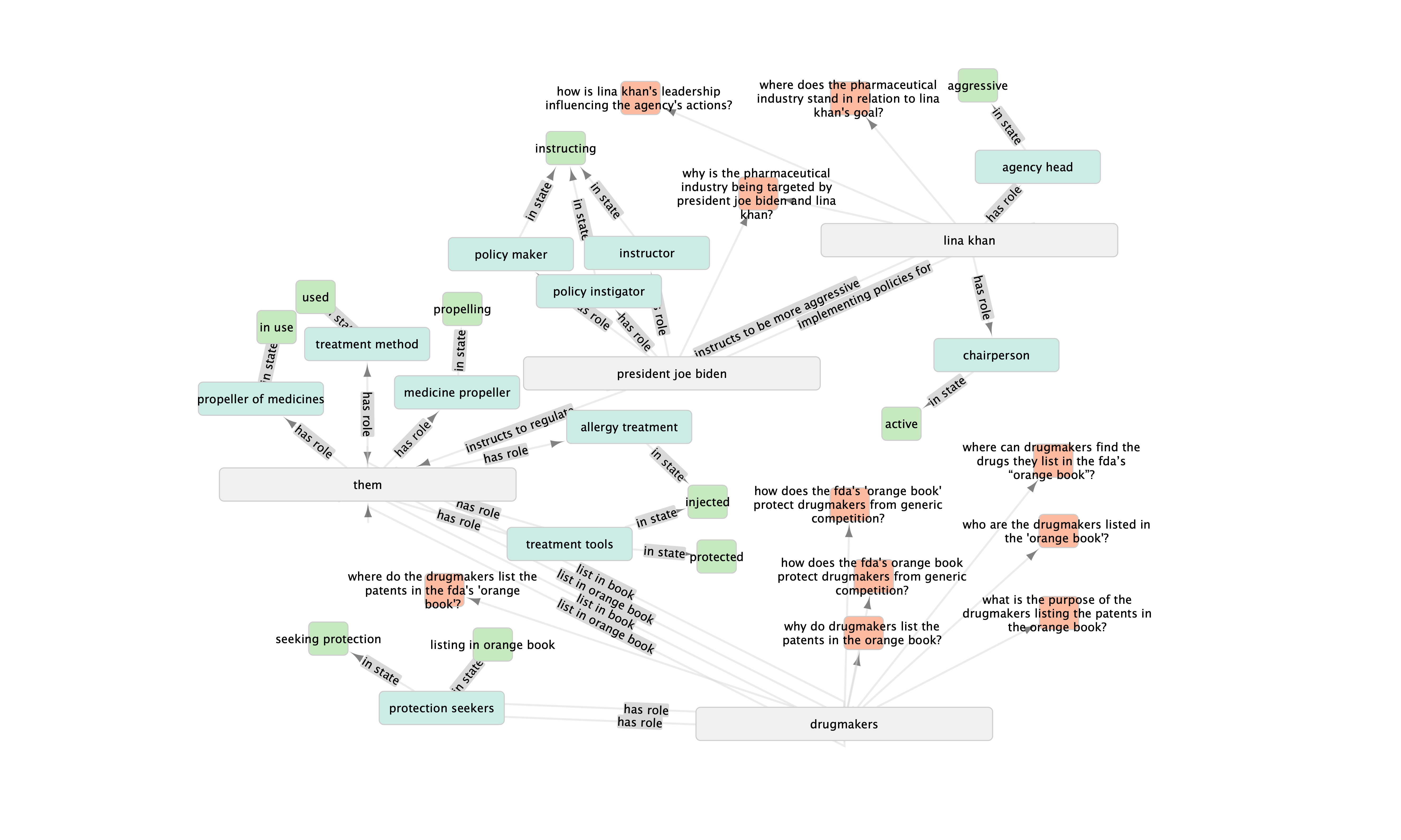}

    \caption{\textbf{Healthcare -} \textit{President Joe Biden has instructed his Federal Trade Commission to be more aggressive in reining in the pharmaceutical industry. Under its chairperson, Lina Khan, the agency is aggressively testing the limits of its powers in pursuit of that goal. The targeted patents cover devices that propel medicines for asthma and emphysema into the lungs or inject epinephrine to treat a severe allergic attack. Drugmakers list them in the FDA’s “Orange Book,” which can afford the products greater protection from generic competition.}}
    \label{fig:hc3}
\end{figure*}

\newpage

\newpage

\begin{figure*}[ht]
    \centering
\includegraphics[trim={20cm 0 20cm 0},clip,width=\textwidth]{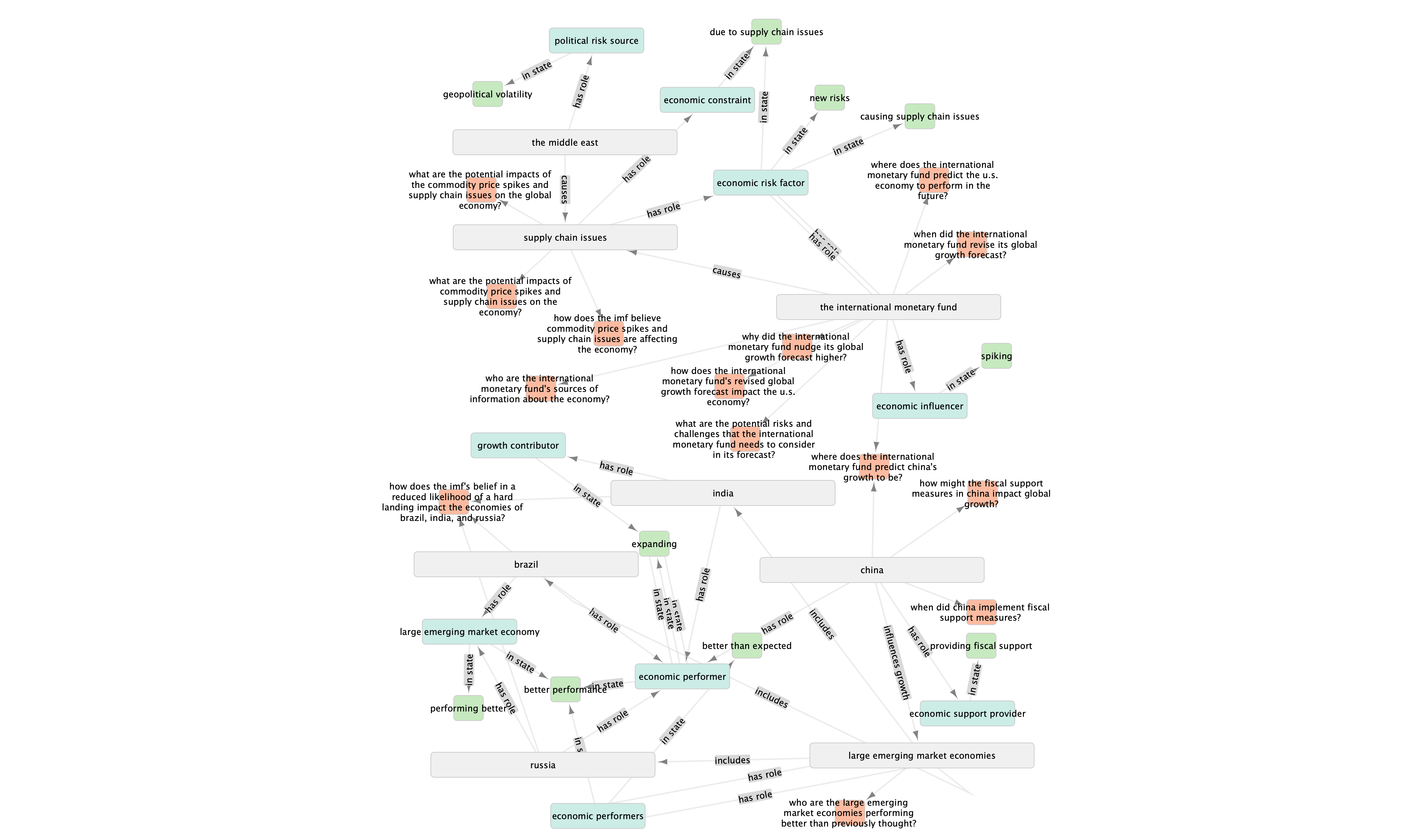}

    \caption{\textbf{Economy -} \textit{The International Monetary Fund on Tuesday nudged its global growth forecast higher, citing the unexpected strength of the U.S. economy and fiscal support measures in China. It now sees global growth in 2024 at 3.1\%, up 0.2 percentage point from its prior October projection, followed by 3.2\% expansion in 2025. Large emerging market economies including Brazil, India and Russia have also performed better than previously thought. The IMF believes there is now a reduced likelihood of a so-called hard landing, an economic contraction following a period of strong growth, despite new risks from commodity price spikes and supply chain issues due to geopolitical volatility in the Middle East.}}
    \label{fig:eco1}
\end{figure*}

\newpage

\begin{figure*}[ht]
    \centering
\includegraphics[trim={10cm 0 10cm 0},clip,width=\textwidth]{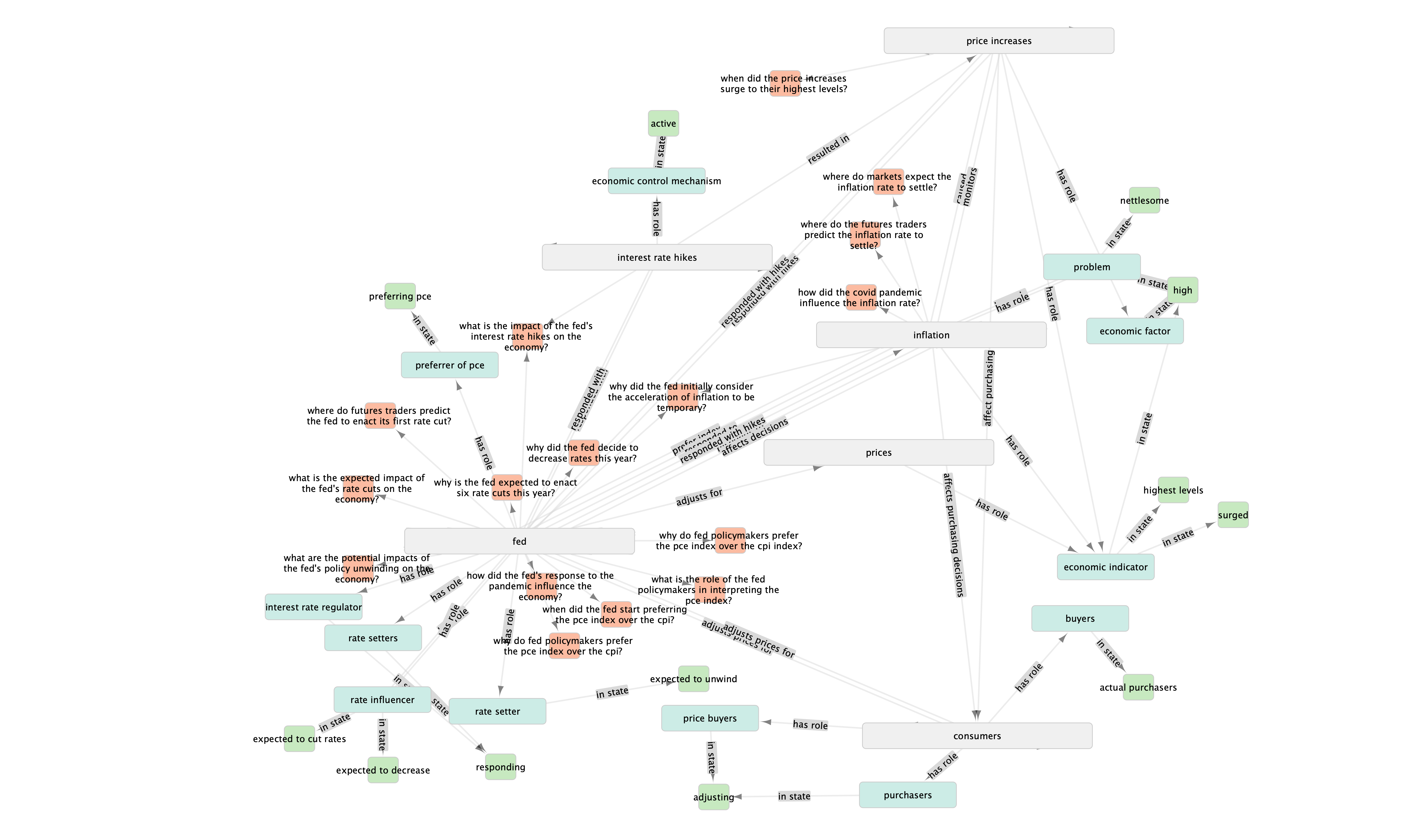}

    \caption{\textbf{Economy -} \textit{While the public more closely follows the Labor Department’s consumer price index, Fed policymakers prefer the PCE because it adjusts for shifts in what consumers actually buy, while the CPI measures prices in the marketplace. Inflation has been a nettlesome problem since the early days of the Covid pandemic, when price increases surged to their highest levels since the early 1980s. The Fed initially expected the acceleration to be temporary, then responded with a series of interest rate hikes that took its benchmark rate to its highest in more than 22 years. Now, with the inflation rate cooling markets largely expect the Fed to start unwinding its policy tightening. As of Friday morning, futures traders were assigning about a 53\% chance the Fed will enact its first rate cut this cycle in March, according to CME Group data. Pricing points to six quarter-percentage point decreases this year.}}
    \label{fig:eco2}
\end{figure*}

\newpage

\newpage

\newpage

\begin{figure*}[ht]
    \centering
\includegraphics[trim={6cm 0 6cm 2cm},clip,width=\textwidth]{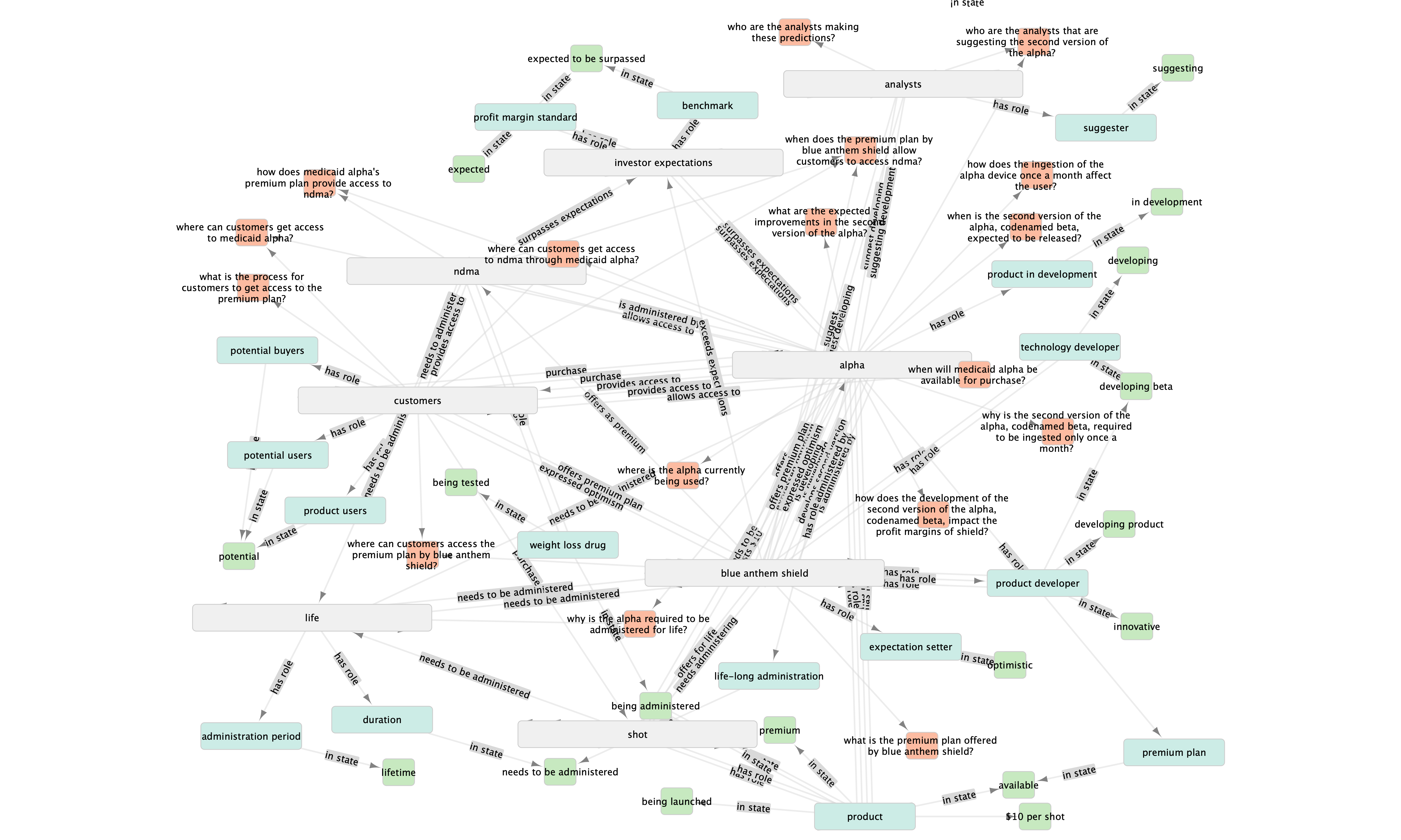}

    \caption{\textbf{Technology Development -} \textit{Medicaid Alpha, the premium plan by Blue Anthem Shield, allows customers to get access to NDMA, the cutting-edge weight loss drug taking the US and UK by storm. Alpha costs \$10 per shot and needs to be administered for life. Shield CEO Lua Li expressed optimism at the annual Shareholder's meeting, stating that the Q2-Q3 profit margins should surpass investor expectations by over 5\%. Analysts suggest Shield is developing a second version of the Alpha, codenamed Beta, that is required to be ingested only once a month.}}
    \label{fig:td1}
\end{figure*}

\newpage

\begin{figure*}[ht]
    \centering
\includegraphics[trim={7cm 0 7cm 0},clip,width=\textwidth]{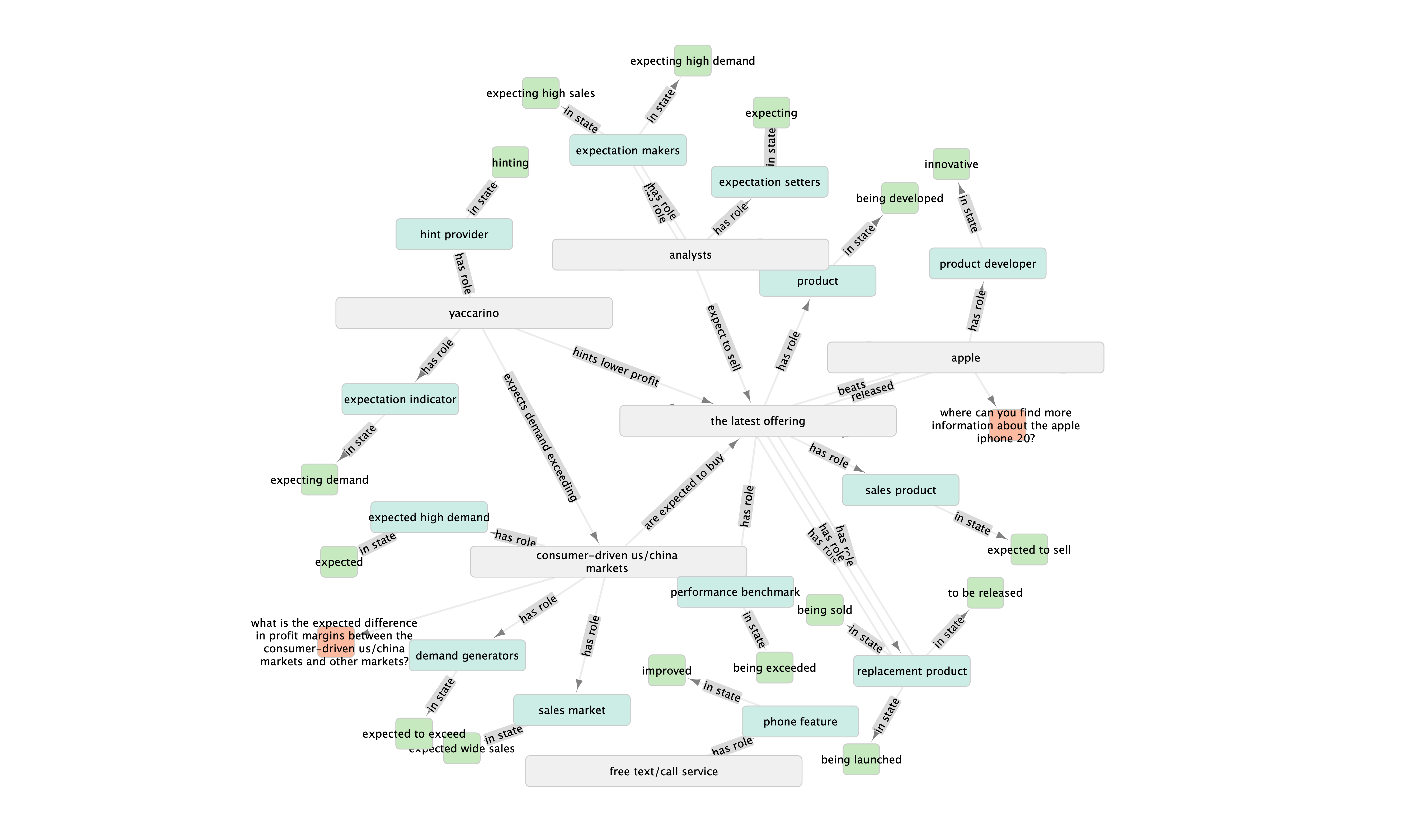}

    \caption{\textbf{Technology Development -} \textit{Apple released its new iPhone 20, the latest phone model rumored to beat market expectations for demand by late 2034. The latest offering boasts a significantly improved battery life of 25 hours, free text/call service, and an all-glass finish. It is also the first phone to be entirely recyclable. The latest phone will replace the old iPhone 19 at the same price of \$999. Analysts expect the phone to sell widely in consumer-driven US/China markets. Yaccarino hinted that the profit margins would be lower on this version, but the demand is expected to exceed that of the 19 by over 20\%.}}
    \label{fig:td2}
\end{figure*}

\newpage

\begin{figure*}[ht]
    \centering
\includegraphics[trim={0cm 0 0cm 0},clip,width=\textwidth]{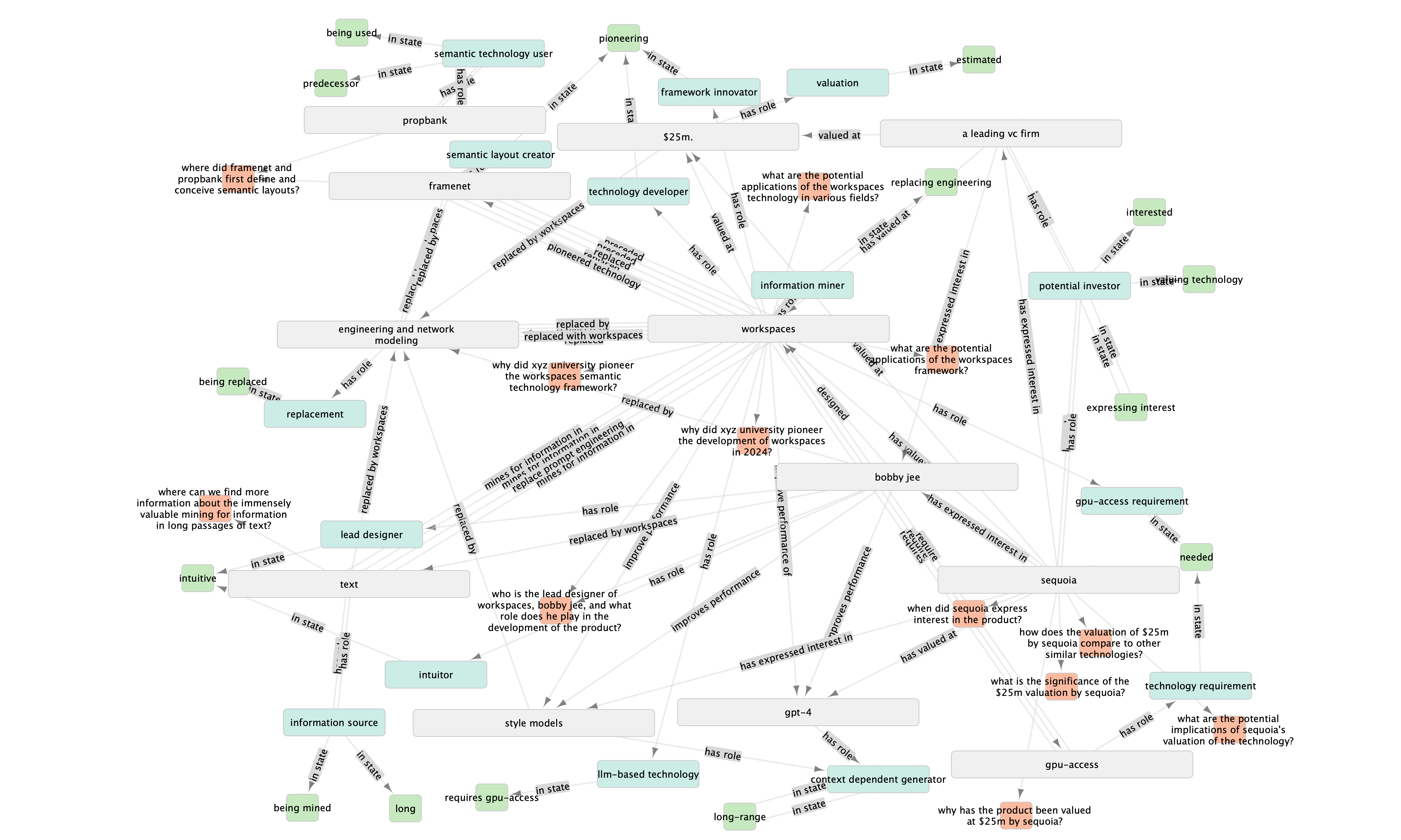}

    \caption{\textbf{Technology Development -} \textit{Workspaces, a novel semantic technology framework pioneered at XYZ university in 2024 promises to revolutionize the semantic layouts defined and conceived first by FrameNet and PropBank in the mid-1900s. Workspaces are an LLM-based technology that requires GPU-access to run, and have proved immensely valuable at mining for information in long passages of text - a replacement to prompt engineering and network modeling. Lead designer of Workspaces, Bobby Jee, intuited that the Workspace models may significantly improve the performance of GPT-4 style models by over 30\% at long-range context dependent generation tasks. Sequoia, a leading VC firm, has expressed interest in the product, and has valued the technology at \$25M.}}
    \label{fig:td3}
\end{figure*}

\end{document}